\documentclass[pdflatex,sn-nature]{sn-jnl}
\usepackage{lineno}

\usepackage{graphicx}%
\usepackage{multirow}%
\usepackage{amsmath,amssymb,amsfonts}%
\usepackage{amsthm}%
\usepackage{mathrsfs}%
\usepackage[title]{appendix}%
\usepackage{xcolor}%
\usepackage{colortbl} 

\usepackage{textcomp}%
\usepackage{manyfoot}%
\usepackage{caption}
\usepackage{booktabs}%
\usepackage{algpseudocode}%
\usepackage{listings}%
\usepackage{subfigure}
\usepackage{multirow}
\usepackage{array} 

\usepackage[table]{xcolor}
\definecolor{lightred}{rgb}{1,0.8,0.8}

\usepackage[ruled,linesnumbered]{algorithm2e}

\usepackage[dvipsnames]{xcolor}


\theoremstyle{thmstyleone}%
\theoremstyle{thmstyletwo}%

\theoremstyle{thmstylethree}%

\definecolor{headergrey}{rgb}{0.88,0.88,0.88} 
\definecolor{rowgrey}{rgb}{0.97,0.97,0.97} 
\definecolor{rowgrey2}{rgb}{0.91,0.91,0.91} 

\unnumbered  

\begin{document}

\title[Physics-informed Diffusion Generative Model for Time-Series Data Synthesis in Dynamic Systems]{Physics-informed Diffusion Generative Model for Time-Series Data Synthesis in Dynamic Systems}

\author[1]{\fnm{Haiteng } \sur{Wang}} 
\email{wanghaiteng@buaa.edu.cn}

\author[4]{\fnm{Yunfei} \sur{Zhu}}\email{zhuyunfei@buaa.edu.cn}

\author[3]{\fnm{Tao} \sur{Wang}}\email{seatao.wang@connect.polyu.hk}

\author[1]{\fnm{Yikang} \sur{Li}}\email{liyikang@buaa.edu.cn}

\author[1]{\fnm{Jiabao} \sur{Dong}}\email{aarondong@buaa.edu.cn}

\author[3]{\fnm{Xiaoge} \sur{Zhang}}\email{xiaoge.zhang@polyu.edu.hk}

\author*[1,2]{\fnm{Lei} \sur{Ren}}\email{renlei@buaa.edu.cn}

\affil*[1]{\orgdiv{School of Automation Science and Electrical Engineering}, \orgname{Beihang University}, \orgaddress{ \city{Beijing}, \postcode{100191}, \country{China}}}

\affil[2]{\orgdiv{Hangzhou International Innovation Institute}, \orgname{Beihang University}, \orgaddress{ \city{Hangzhou}, \postcode{311115}, \country{China}}}

\affil[3]{\orgdiv{Department of Industrial and Systems Engineering}, \orgname{The Hong Kong Polytechnic University}, \orgaddress{ \city{Kowloon}, \postcode{100191}, \country{Hong Kong}}}

\affil[4]{\orgdiv{School of Software}, \orgname{Beihang University}, \orgaddress{ \city{Beijing}, \postcode{100191}, \country{China}}}


\abstract{
Industrial time-series signals, such as turbine temperature and rotational speed in aero-engines, are essential for monitoring the health and operational status of complex dynamical systems.
However, collecting such data is often limited by harsh environments (e.g., high temperature and high pressure) and the high cost of experimental testing.
To address this challenge, we introduce PhysDGM, a stepwise physics-embedded diffusion generative model for synthesizing time-series data that are consistent with the underlying physical laws of dynamical systems. PhysDGM embeds physical laws directly into each reverse diffusion step of the generative process, ensuring trajectory-level physical consistency, rather than enforcing constraints only at the final output. A large-scale AI-synthetic dataset (4.4 million samples, 20× scale-up) constructed by PhysDGM demonstrates strong fidelity across 34 datasets spanning turbofan engines, aero-engines, batteries, and chemical processes. After incorporating the synthetic data, the downstream task performance substantially surpassed that using real data alone by 48\% for remaining useful life prediction, 15\% for health indicator estimation, 22\% for state-of-health assessment, and 20\% for fault diagnosis. Moreover, it requires 10–20× less training data than existing approaches, substantially reducing the high cost of data collection in dynamical systems. We further demonstrate PhysDGM’s potential in identifying early-stage faults in aero-engines by incorporating AI-synthesized data. In summary, PhysDGM provides a solid foundation for generating physically consistent industrial time-series, paving the way for expanding physics-guided AI into diverse data-scarce environments, including both industrial machinery and complex chemical reaction dynamics.

}




\maketitle

\newpage

\section{Introduction}

Large-scale, high-quality time-series data forms the foundation~\cite{kusiak2017smart} for driving artificial intelligence (AI) in dynamic system modeling and industrial applications. However, collecting data for systems such as aero-engines is significantly hindered by harsh environments (e.g., high temperature and high pressure) and the prohibitive testing cost~\cite{li2024learning, yin2017location}. The scarcity of high-quality training data limits the reliability and deployability of advanced AI models in critical applications such as industrial automation~\cite{ng2020predicting, wang2024physics, qu2022controlling, lee2025active} and process monitoring~\cite{jiang2021data, bitharas2022interplay, liu2023lightweight, zheng2023interval}. To address this challenge, data augmentation~\cite{chawla2002smote, guo2004learning,pourhabib2015absent} and synthetic data generation~\cite{chen2025versatile, ren2025aigc, pan2023vae, lin2020anomaly, Che2024AdaptiveMultiHead, He2023AttributerelevantDistributeda, ren2024diffmts, yang2023ddmt,kong2022diffwave,lopezalcaraz2022diffusionbased,fei2022towards,qiao2024state} have emerged as promising solutions. Recent advances in diffusion probabilistic models for text-to-image synthesis, such as Stable Diffusion~\cite{esser2024scaling} and DALLE 3~\cite{betker2023improving}, have demonstrated unprecedented fidelity in modeling complex, high-dimensional data characteristics. Analogously, the diffusion-based data generation paradigm~\cite{ren2024diffmts,wang2025metaindux,eivazi2024diffbatt,luo2023state} has been adapted to synthesize time-series data through specialized architectures such as Diff-MTS~\cite{ren2024diffmts}, DiffWave~\cite{kong2022diffwave}, and TabDDPM~\cite{kotelnikov2023tabddpm}, which significantly outperform traditional generative adversarial networks or variational autoencoders in capturing temporal dependencies and long-range correlations.

However, most generative models remain fundamentally physics-blind when applied to dynamical systems. That is, they treat time-series generation as a purely data-driven task, ignoring the underlying physical laws (e.g., thermodynamics, turbulent dynamics). If the synthetic data produced by generative models does not respect underlying physical principles of dynamic systems, it may mislead the model into learning nonphysical and unrealistic behavior, potentially compromising the utility and reliability of the model in the downstream tasks.

To mitigate the limitations of purely data-driven approaches, physics-informed neural networks (PINNs)\cite{karniadakis2021physics,lu2021learning,Wei2025,rao2023encoding,patki2016synthetic,seo2021controlling,xiong2023controlled,azizzadenesheli2024neural, li2024physics} have emerged as a powerful paradigm to incorporate physical knowledge into the learning process. In recent years, representative studies have explicitly encoded physical laws into neural network~\citep{rao2023encoding,patki2016synthetic,seo2021controlling}. For example, PeRCNN~\citep{rao2023encoding} forcibly encodes known partial differential equation structures to model complex spatiotemporal dynamics like reaction–diffusion processes and achieves significant improvements in physical consistency. However, this paradigm is ill-suited for time-series generative tasks, as rigid architectural constraints often over-constrain the solution space, preventing the synthesis of diverse and representative data necessary for robust downstream training. Alternatively, soft constraints in the loss function of neural network are often preferred due to their ease of implementation~\cite{xiong2023controlled,azizzadenesheli2024neural, li2024physics}. These approaches typically operate by penalizing the deviation of neural network's output from well-established physical laws in the form of PDE and its variants, treating the model as a single-step mapping function from input to constrained output. However, this final-stage enforcement strategy becomes insufficient for generative models of iterative nature, such as diffusion models, which synthesize data through a multi-step refinement process that simulates physical diffusion dynamics~\cite{song2021score,ho2020denoising}. If physical consistency is imposed solely at the model output, deviations of intermediate model outputs from physical laws will accumulate and propagate. This error accumulation irreversibly drives the generation process away from the physically valid manifold, yielding invalid data even if the final outputs are forced to comply with the physical constraints.

Here, we argue that to synthesize time-series data for dynamical systems, physical laws should be treated not as post-hoc penalties applied at the final stage, but as an active feedback mechanism throughout the whole generation process. Drawing inspiration from control theory, we treat model generation as a closed-loop system and use synchronous feedback from the physical governing equations to steer rollouts toward physically meaningful solution manifold at each refinement step.

We implement this vision through the stepwise Physics-embedded Diffusion Generative Model (PhysDGM), a framework that transforms physical laws from passive constraints into active multi-step guidance signals to shape its generation behavior in physical compliance. By integrating a stepwise physical embedding mechanism, PhysDGM iteratively steers the generative trajectory toward physically valid manifolds at each diffusion step, effectively preventing the accumulation of errors inherent in unconstrained diffusion. Moreover, a gradient-guided physical sampling scheme is proposed to transform constraint penalties into gradient guidance, enabling the model to flexibly adapt to evolving physical environments without retraining. Finally, we validate the universality of this approach across 34 diverse industrial datasets, ranging from aero-engine degradation to chemical process fault diagnosis. Computational results suggest that PhysDGM not only consistently enhances downstream predictive accuracy by an average of 15\%–48\% but also significantly lowers the barrier for deploying AI under changing environments, achieving high data fidelity with 10–20 times less training data than several advanced generative models in the literature.

\begin{figure*}[!htpb]
	\centerline{\includegraphics[width=0.99\linewidth]{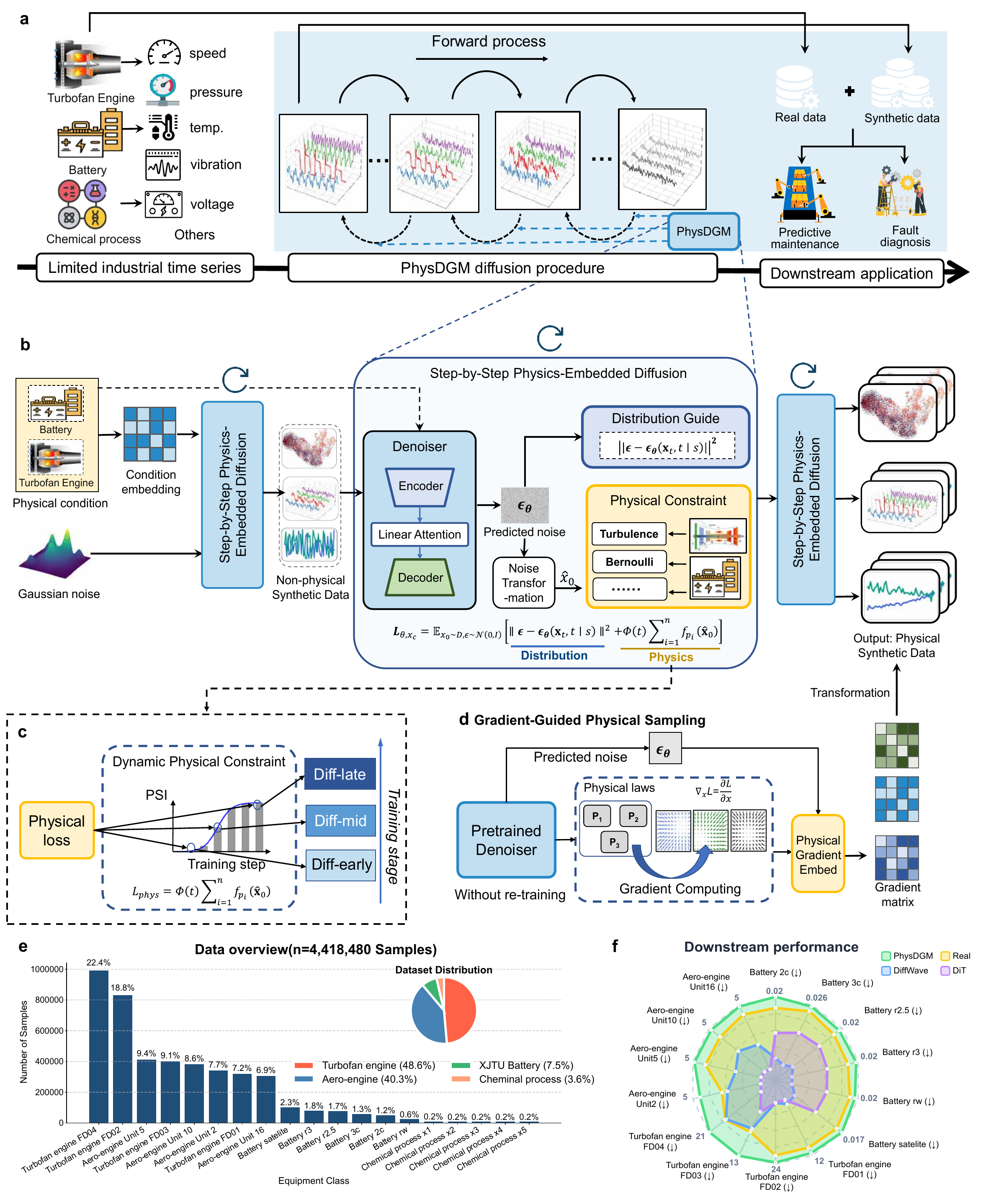}}
	\caption{\textbf{Overview of PhysDGM}. \textbf{a}, Schematic of PhysDGM’s data generation process.
The forward process gradually perturbs structured time-series data through a predefined diffusion schedule and PhysDGM generates structured time-series data from random noise by progressively denoising through a reverse process guided by embedded physical constraints. 
\textbf{b}, The framework integrates physics-embedded diffusion with  distribution guidance, incorporating dynamic physical constraints and physical laws (e.g., degradation, trends) to ensure physically plausible outputs. 
\textbf{c}, Dynamic physical constraints are progressively applied across training stages (Diff-early, Diff-mid, Diff-late). 
\textbf{d}, Gradient-guided sampling refines generated data by embedding physical laws through gradient adjustments, leveraging a pretrained denoiser without retraining. 
\textbf{e}, Synthetic equipment dataset distribution of $\approx 4.4$ million records across different industrial equipment types. Detailed descriptions of the specific datasets and their sub-categories are provided in Supplementary Data Table~\ref{tab:C-MAPSS}-~\ref{tab:TEP_Disturbances}. 
\textbf{f}, Performance improvement via data augmentation. Only 5\% of real training data with PhysDGM-generated samples (dark bars) significantly reduces RMSE across diverse downstream datasets. }
	\label{fig:框架示意图}
\end{figure*}

\section{Results}

\subsection{Overview of PhysDGM}
Data scarcity fundamentally bottlenecks the modeling and prediction of industrial dynamical system states. The critical task is to synthesize large-scale, physically consistent time-series data from sparse observations.

We introduce PhysDGM\textemdash a stepwise physics-embedded diffusion generative model that incorporates physical constraints throughout the entire data generation process (Fig.~\ref{fig:框架示意图}). As only a limited set of real-world time-series data samples are available in many industrial applications (e.g., turbofan engines or chemical processes), our PhysDGM reverse diffusion process takes a progressive approach to synthesize high-fidelity synthetic data step by step (Fig.~\ref{fig:框架示意图}a). At a high level, PhysDGM generates physically valid synthetic data through three key innovations. First, its stepwise physical embedding mechanism progressively enforces physical laws at each reverse diffusion step, providing fine-grained guidance signal and ensuring compliance to physical laws throughout the entire data generation process (Fig.~\ref{fig:框架示意图}b). Second, PhysDGM’s dynamic physical constraint training strategy cleverly resolves the potential conflict between data distribution learning and physical consistency.  It achieves this by prioritizing likelihood fitting in the early stage of model training and progressively increasing the weight of physics regularization (Fig.~\ref{fig:框架示意图}c). Finally, we design a gradient-guided physical sampling method that converts the violations error of physical equations into guidance gradients, enables fast adaptation to new physical environments without model retraining (Fig.~\ref{fig:框架示意图}d). 

Our PhysDGM framework was comprehensively validated across 34 industrial datasets of heterogeneous tasks spanning diverse domains including turbofan engines, aero-engines, batteries, and chemical processes. The downstream tasks comprised RUL prediction, HI prediction, SOH estimation, and fault diagnosis. As detailed in Fig.~\ref{fig:框架示意图}e, PhysDGM generated a large-scale, multi-source dataset of approximately 4.4 million synthetic samples to augment model training. Crucially, after incorporating data generated by PhysDGM (Fig.~\ref{fig:框架示意图}f), the downstream models achieved lower $\text{RMSE}$ than those trained solely on real data, outperforming advanced generative baselines such as DiT and DiffWave. These results underscore a significant breakthrough: PhysDGM produces high-fidelity synthetic data that enables models to exceed the predictive performance attainable with real datasets alone, representing a shift toward physics-grounded Generative AI for creating ideal datasets.

\subsection{Results Overview}

PhysDGM was trained on diverse real-world time-series datasets from industrial dynamical systems and then utilized to generate synthetic datasets scaled from 1× to 20× the size of the original dataset. The synthetic data were validated across multiple downstream tasks, including RUL prediction, SOH estimation, HI prediction, and fault diagnosis.

To comprehensively assess PhysDGM’s capabilities, we conducted a multi-faceted evaluation:
(1) the distribution characteristics of the synthetic data were assessed across 34 datasets spanning aero-engines, turbofan engines, batteries, and chemical processes;
(2) the physical fidelity of the synthetic data was evaluated quantitatively and qualitatively across multiple representative physical constraints;
(3) the model performance was evaluated across diverse downstream tasks, including RUL prediction, HI prediction, SOH estimation, and fault diagnosis in few-shot settings.
(4) the cross-model generalization of synthetic data was verified across several representative deep learning architectures.


\subsubsection{PhysDGM achieves high-fidelity synthetic data}

\begin{figure*}[htbp]  
    \centering  
    \subfigure{\includegraphics[width=1.0\linewidth]{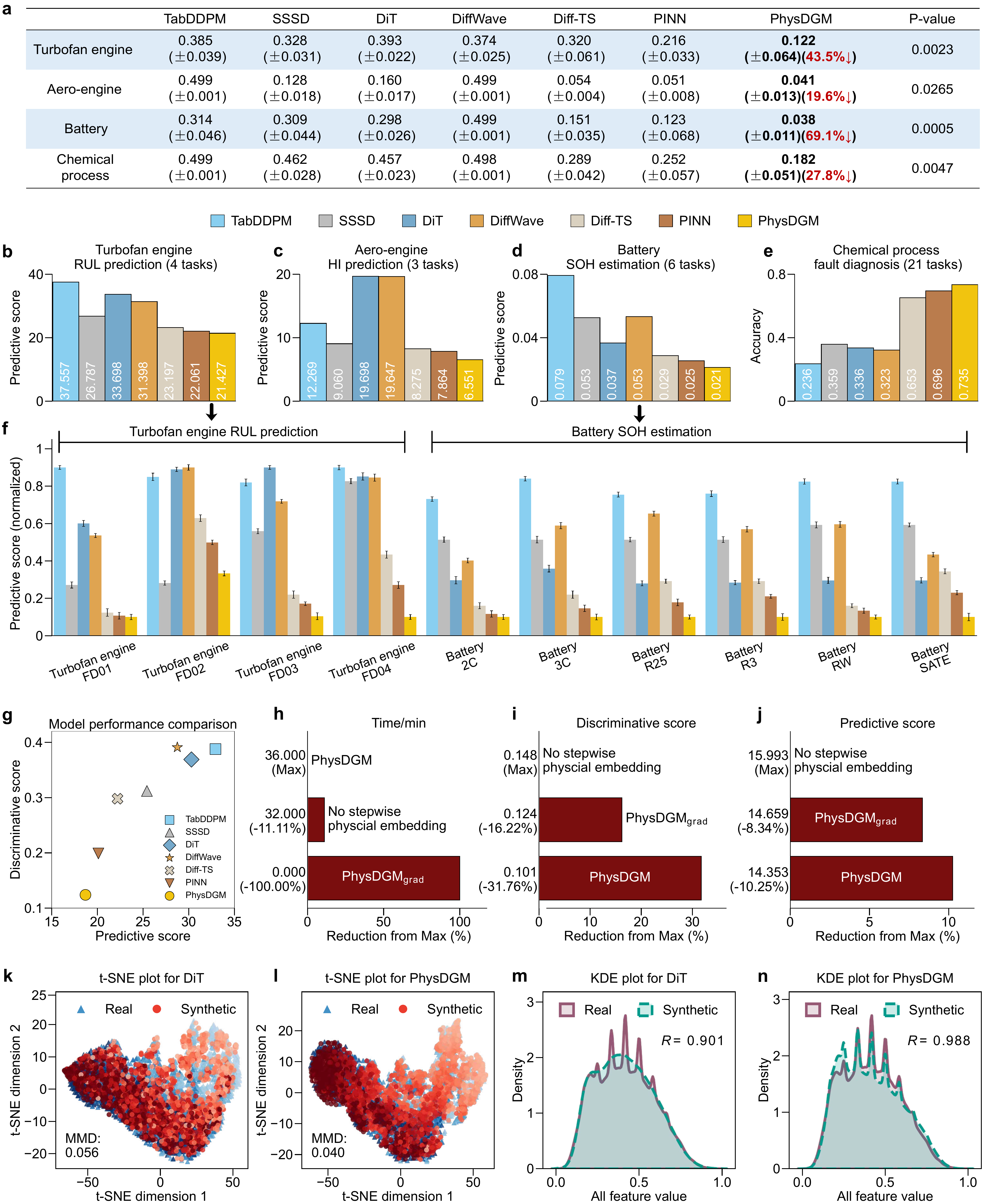}}
    
    \caption{
        \textbf{Quality inspection of data generated by PhysDGM.} 
        \textbf{a}, Fidelity comparison of data generated by PhysDGM and other models. One-sided Wilcoxon signed-rank test is utilized to calculate $p$ values.
        \textbf{b.c.d.e}, Average performance of the utility test for data generated by PhysDGM and other models across various tasks. The assessment quantifies the performance of synthetic-trained models on real test data with RMSE for regression tasks and accuracy for classification tasks.
        \textbf{f}, Sub-task performance of the utility test for data generated by PhysDGM and other models. The mean values and standard deviations shown in a--f are derived from 7 independently trained models, each evaluated over 1000 experimental trials.
        \textbf{g}, Comparison of the average discriminative and predictive scores for data generated by PhysDGM and other models. These predictive scores are obtained by training downstream models (transformers or classifiers) solely on synthetic data and evaluating their RMSE or accuracy on the original test set.
        \textbf{h.i.j}, Comparison of discriminative and predictive scores for data generated of $\text{PhysDGM}$ , $\text{PhysDGM}_{\text{grad}}$, and $\text{PhysDGM}_{\text{non-phys}}$.
        \textbf{k.l.m.n}, Visualization of PhysDGM-generated data through t-SNE analysis, and Kernel Density Estimation (KDE). These visualizations compare the distributional overlap of original versus synthetic data using t-SNE embeddings and their probability density curves on key features. 
    }
    \label{fig:figure2}  
\end{figure*}

PhysDGM is designed to generate high-quality synthetic time series that preserve the physics characteristics of the original data and support effective training of downstream models. We evaluate the effectiveness of PhysDGM on four benchmark tasks (Fig.~\ref{fig:figure2}): turbofan engine RUL prediction, aero-engine HI prediction, battery SOH estimation, and chemical-process fault diagnosis (See \textbf{Methods} section \textbf{Dataset Description}) against multiple state-of-the-art models, including PINN~\cite{wang2024physics} and diffusion models such as DiffWave~\cite{kong2022diffwave}, SSSD~\cite{lopezalcaraz2022diffusionbased}, TabDDPM~\cite{kotelnikov2023tabddpm}, DiT~\cite{peebles2023scalable}, and Diff-TS~\cite{yuandiffusion}. PINN generates data by optimizing neural networks under physical constraints in the form of differential equations. In contrast, diffusion models generally operate by corrupting data with noise and then reversing this process to reconstruct the original data.

First, we conducted a discriminative test to evaluate data fidelity. The discriminative score measures the distinguishability between the synthetically generated and real-world data, where a value closer to $0$ indicates greater similarity (See \textbf{Methods} section \textbf{Quality Evaluations Metrics}). As shown in Fig.~\ref{fig:figure2}a, PhysDGM achieved scores 43.5\%, 19.6\%, 69.1\% and 27.8\% lower (i.e., better) than the best-performing baseline on each of the four datasets, respectively. These results confirm that PhysDGM generates data with demonstrably higher fidelity than all competing models. Beyond data fidelity, we next assessed the utility of the synthetic data in training  predictive models for several downstream tasks. PhysDGM outperformed the best-performing baselines by a margin of 7.6\%, 20.8\%, 27.5\%, and 5.6\% across the four tasks (Fig.~\ref{fig:figure2}b-e) and consistently led across all sub-datasets Fig.~\ref{fig:figure2}g; Extended Data Fig.~\ref{fig:exfig1}a,b). Joint evaluation of these metrics (Fig.~\ref{fig:figure2}g) suggests that PhysDGM significantly surpasses all baseline models in both fidelity and utility, demonstrating superior overall data quality.

To adapt to new physical conditions without costly retraining, we propose $\text{PhysDGM}_{\text{grad}}$, a training-free variant that employs a gradient-guided physical sampling strategy. By computing gradients from predefined physical losses, $\text{PhysDGM}_{\text{grad}}$ steers the denoising trajectory along the steepest-descent direction, explicitly reducing physical inconsistencies at each inference step without additional training. We benchmarked the fine-tuning time, discriminative score, and predictive score of $\text{PhysDGM}_{\text{grad}}$ against two counterparts: the standard \text{PhysDGM} (which integrates constraints via stepwise physical embedding during training) and an unconstrained baseline $\text{PhysDGM}_{\text{non-phys}}$ (Fig.~\ref{fig:figure2}h–j). Although PhysDGM achieves the highest fidelity, evidenced by the largest reduction in discriminative and predictive scores, it requires a fixed training period. In contrast, $\text{PhysDGM}_{\text{grad}}$ completely eliminates the need for model retraining (Fig.~\ref{fig:figure2}h), offering a pragmatic trade-off between performance and efficiency. While $\text{PhysDGM}_{\text{grad}}$ incurs a modest performance drop compared to the fully trained $\text{PhysDGM}$, it significantly outperforms the unconstrained baseline. These results establish $\text{PhysDGM}$ as the gold standard for high-fidelity synthetic data generation, while demonstrating that our gradient-guided strategy $\text{PhysDGM}_{\text{grad}}$ provides a robust, training-free mechanism for rapidly adapting the model to dynamic physical constraints.

Finally, we visually substantiated the distributional similarity between generated and real data using t-SNE visualizations (Fig.~\ref{fig:figure2}k,l) and kernel density estimation (KDE) plots (Fig.~\ref{fig:figure2}m,n). We quantified the manifold overlap observed in t-SNE using the Maximum Mean Discrepancy (MMD) and validated the feature density alignment in KDE using the Pearson correlation coefficient ($R$).
In turbofan engine RUL prediction task, PhysDGM achieved a higher distributional overlap with a lower MMD (0.040) compared to the DiT baseline (0.056) (See Extended Data Fig.~\ref{fig:exfig1} for evaluation based on the aero-engine and chemical process datasets). Similarly, the KDE analysis (Fig.~\ref{fig:figure2}m,n) reveals nearly identical density curves of PhyDGM with $R$-value of 0.988 (surpassing DiT's 0.901). This combined visual and quantitative superiority confirms that PhysDGM accurately captures the complex underlying data characteristics, thereby establishing its high fidelity.

\begin{figure*}[!ht]  
    \centering
		\includegraphics[width=1.0\linewidth]{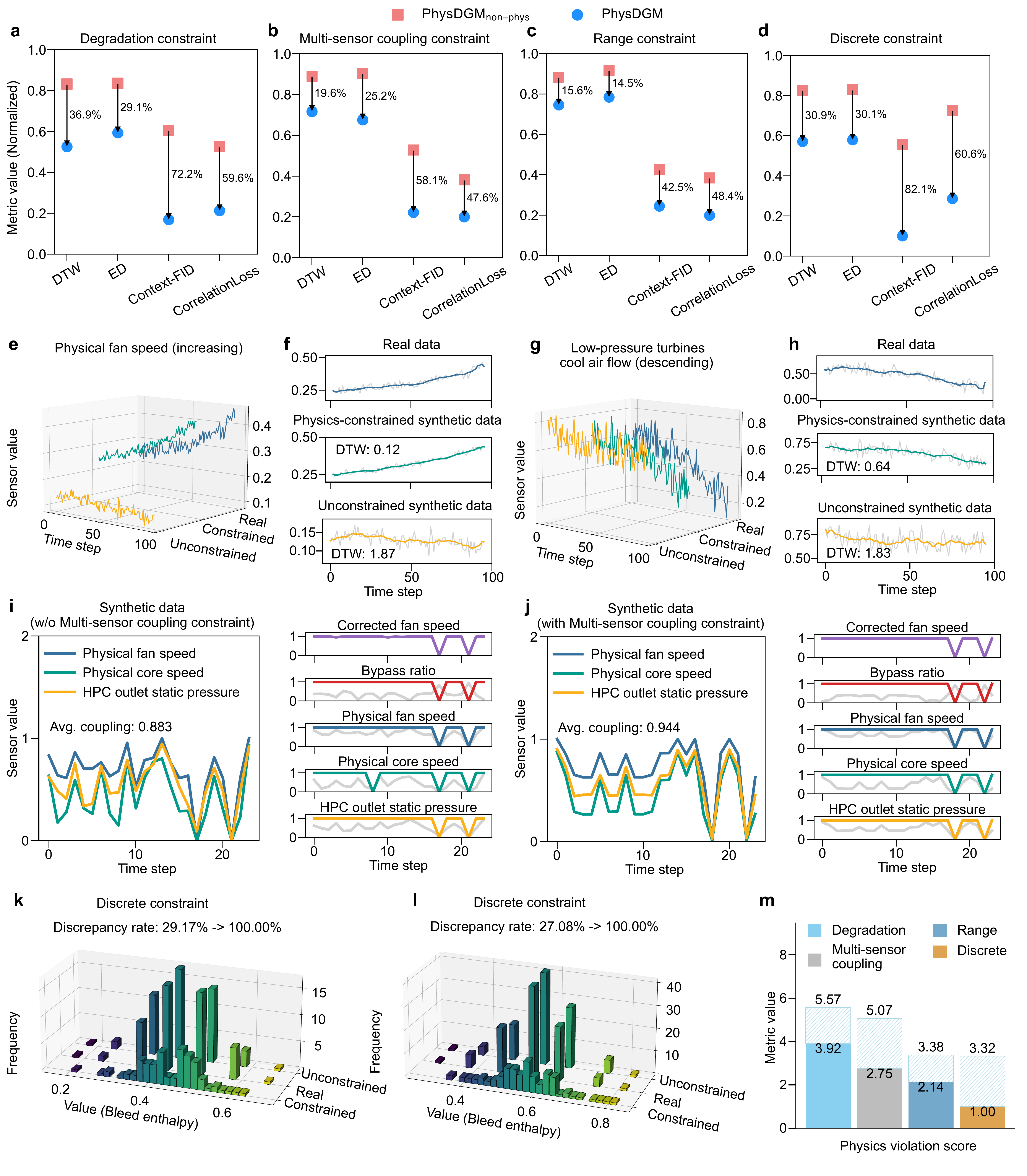} 
   \caption{
        \textbf{Validation of physically consistent synthetic data generation.}
        \textbf{a-d}, Quantitative evaluation of generation fidelity across four physical constraint types: degradation (a), multi-sensor coupling (b), range (c), and discrete (d) constraints. $\text{PhysDGM}$ consistently outperforms the unconstrained baseline $\text{PhysDGM}_{\text{non-phys}}$ across normalized metrics (DTW, ED, Context-FID, and Correlation Loss). Percentages indicate the relative improvement in metric values. 
        \textbf{e-h}, Visual assessment of degradation constraints. 3D and 2D temporal profiles of variables with monotonic trends: physical fan speed (increasing, e, f) and low-pressure turbine cool air flow (descending, g, h). 
        \textbf{i,j}, Impact of multi-sensor coupling constraints. Synthetic data generated without (i) and with (j) coupling constraints.
        \textbf{k,l}, Effectiveness of discrete constraints. Frequency distributions for discrete variables (e.g., Bleed enthalpy) show the correction of invalid continuous values (unconstrained) to valid discrete sets (constrained) with improved discrepancy rates.
        \textbf{m}, Aggregate physics violation scores. Comparison of physics violation score for synthetic data generated by PhysDGM and the unconstrained model.}
    \label{fig:figure3}  
\end{figure*}

\subsubsection{PhysDGM generates physically-consistent synthetic data}

To validate the preservation of explicit physical constraints in the synthetically generated data, we performed an ablation study comparing $\text{PhysDGM}$ against the unconstrained baseline ($\text{PhysDGM}_{\text{non-phys}}$). We evaluated four distinct constraint types: degradation constraints, coupling constraints, fixed range constraints, and discrete constraints. Performance was assessed using Dynamic Time Warping (DTW) for measuring sequence similarity, Euclidean Distance (ED) to quantify dissimilarity at corresponding time steps, ContextFID for statistical fidelity, and Correlation Loss for examining temporal dependencies. PhysDGM demonstrated substantial improvements over $\text{PhysDGM}_{\text{non-phys}}$, reducing error metrics by an average of 25.8\% (DTW), 24.7\% (ED), 84.1\% (ContextFID), and 73.3\% (Correlation Loss) (Fig.~\ref{fig:figure3}a–d) across four constraints. These results indicate that incorporating physical constraints significantly enhances the structural and statistical similarity of synthetic data to the original data.

In addition, we examined the turbofan engine degradation dataset to elucidate the mechanisms underlying these quantitative gains. As this dataset contains well-defined physical principles such as thermodynamic laws and mechanical degradation patterns, we stratified the variables based on three distinct physical characteristics. First, for variables exhibiting strong monotonicity (e.g., physical fan speed, low-pressure turbine cool air flow), PhysDGM successfully recovered the degradation trends, whereas the unconstrained model diverged significantly (Fig.~\ref{fig:figure3}e,f). This degradation trend alignment was quantitatively supported by a reduction in DTW distance of 93\% and 65\%, respectively. Second, for strongly coupled variables (e.g., bypass ratio and physical fan speed), the unconstrained baseline failed to capture inter-variable dependencies; conversely, the application of coupling constraints effectively restored these physical relationships (Fig.~\ref{fig:figure3}g,h). Third, for variables with inherent discreteness (e.g., Bleed Enthalpy), discrete constraints ensured that generated values remained strictly within valid discrete sets. In contrast, the unconstrained method yielded invalid, continuous distributions that failed to reproduce the underlying data structure (Fig.~\ref{fig:figure3}i,j). 

Beyond general structural similarity, we quantified the precise adherence to physical laws using constraint-specific violation metrics (See \textbf{Method} section \textbf{Embedded Physical Constraints} for details). The integration of physical constraints significantly mitigated physical inconsistencies, reducing violation metrics by 29.7\%, 45.6\%, 36.7\%, and 70.0\%, respectively (Fig.~\ref{fig:figure3}k). These results confirm that PhysDGM substantially enhances the physical realism of the generated data.

\subsubsection{Stepwise physical embedding significantly improves data quality}

We assessed the efficacy of the stepwise physical embedding strategy by benchmarking it against a traditional PINN approach that incorporates physical constraints in the loss function of the neural network. Specifically, we trained separate transformer models using synthetic data generated from each method and compared their RMSE on the original test sets (Extended Data Fig.~\ref{fig:exfig2}a-d). The results demonstrate that the data generated by PhysDGM shows significantly higher utility, reducing RMSE by 59.7\%, 14.6\%, 50.0\%, and 12.6\% across four datasets compared to the PINN baseline. This finding strongly suggests that our progressive constraint scheme achieves a superior balance between capturing complex data distributions and adhering to physical laws, thereby producing synthetic data with substantially higher physical significance.

We further investigated the influence of the decay parameter $\alpha$ on the stepwise mechanism (Extended Data Fig.~\ref{fig:exfig2}e-h). A hyperparameter sweep across $\alpha \in \{1, \dots, 5\}$ identified $\alpha = 3$ as the optimal setting, offering the most balanced trade-off between learning flexibility and physical consistency. Moreover, PhysDGM exhibited significant robustness, maintaining high generation fidelity even under suboptimal $\alpha$ values.

\begin{figure*}[!ht]  
    \centering  
    \hspace*{-0.5cm} 
    \includegraphics[width=1.0\linewidth]{fig/main_fig/fig4_v6_edited.pdf}
    \captionsetup{labelformat=empty}
    \caption{}
    \label{fig:figure4} 
\end{figure*}
\addtocounter{figure}{-1}
\begin{figure*} [t!]
    \caption{\textbf{Performance of PhysDGM's generated data on downstream tasks.} \textbf{a}, Performance comparison of models on specified test sets using mixed datasets, where synthetic data generated with varying synthetic ratios is combined with the original data for RUL prediction. 
    \textbf{b}, Performance comparison of data generated by PhysDGM and other models on turbofan engine RUL prediction across various evaluation metrics.
    \textbf{c.d}, Performance of data generated by PhysDGM in the task of RUL prediction.
    \textbf{e}, Performance comparison of models on specified test sets using mixed datasets, where synthetic data generated with varying synthetic ratios is combined with the original data for aero-engine HI prediction. 
    \textbf{f}, Performance comparison of data generated by PhysDGM and other models on aero-engine HI prediction across various evaluation metrics
    \textbf{g}, Performance of data generated by PhysDGM in the task of aero-engine HI prediction.
    \textbf{h}, Performance comparison of models on specified test sets using mixed datasets, where synthetic data generated with varying synthetic ratios is combined with the original data for battery SOH estimation. 
    \textbf{i}, Performance comparison of data generated by PhysDGM and other models on battery SOH estimation across various evaluation metrics
    \textbf{j.k}, Performance results of PhysDGM-generated data on the state of health (SOH) estimation task. 
    \textbf{i}, Performance of PhysDGM on the chemical process fault diagnosis task, where models are trained on mixed datasets combining original data with synthetic data generated by different models and then evaluated on the test set.
    \textbf{m.n}, AUROC curves for the fault diagnosis task using the original data versus the original data combined with PhysDGM's generated data.
    \textbf{o.p.q.r}, Performance of PhysDGM on the turbofan engine RUL prediction task under a 5\% small sample setting.
    }  
    \label{fig:figure4}  
\end{figure*}

\subsubsection{PhysDGM improves equipment health prognostics}

We validated the practical engineering utility of PhysDGM by applying it to three critical predictive maintenance tasks: turbofan engine RUL prediction, aero-engine HI prediction, and battery SOH estimation. In these tasks, we augmented original training data with PhysDGM-generated synthetic samples to train downstream prediction models at ratios ranging from 100\% (1x) to 2000\% (20x). Performance was systematically compared against models trained only with the original data and those augmented by samples generated by advanced generative models, such as DiffWave and DiT.

As the volume of PhysDGM's synthetic data increased, the RMSE for all prediction tasks showed a steady downward trend (Fig.~\ref{fig:figure4}a,e,h). Specifically, increasing the augmentation scale from 1x to 20x further reduced RMSE by 21.0\%, 12.2\%, and 11.1\% for the RUL, HI, and SOH tasks, respectively. More importantly, compared to models trained solely on raw data, the 20x augmentation strategy reduced the RMSE for three prediction tasks by 47.6\%, 15.4\%, and 21.5\%. Conversely, competing models such as DiT exhibited performance degradation at higher augmentation volumes. This observation suggests that scaling up low-fidelity synthetic data, which often lack physical plausibility, introduces noise or spurious patterns that mislead downstream models. In contrast, PhysDGM outperforms all baselines, indicating that it generates high-fidelity samples that effectively expand the training manifold and enhance predictive accuracy. To verify the robustness of these improvements, we evaluated synthetic data generalization of PhysDGM across diverse network architectures including CNN, LSTM, transformer, TLSTM, and MCTAN. Experiments show consistent performance gains across multiple evaluation metrics such as RMSE, $R^2$, and MAE. (Extended Data Fig.~\ref{fig:exfig3}). In addition, we conducted a comprehensive multi-dimensional evaluation including MAD, $R^2$, MAE, MAPE, RMSPE, to assess the utility of the generated data for downstream tasks. The results demonstrate that PhysDGM-augmented models exhibit optimal performance across all evaluation dimensions (Fig.~\ref{fig:figure4}b,f,i). Additionally, visual comparisons of predicted versus actual values for the RUL prediction (Fig.~\ref{fig:figure4}c,d,g) and SOH estimation (Fig.~\ref{fig:figure4}j,k) tasks reveal exceptionally high consistency and alignment. This robust predictive fidelity indicates that models trained on PhysDGM-generated data possess the potential for direct deployment in real-world predictive maintenance systems, capable of providing reliable decision support for the health management of critical industrial equipment.

\subsubsection{PhysDGM improves fault diagnosis in chemical processes}

To systematically evaluate the effectiveness of PhysDGM for chemical process fault diagnosis, we used a benchmark chemical fault diagnosis dataset and compared models trained on real data alone with those trained on 1:1 mixtures of real data and synthetic data from multiple generators. Performance was assessed using two widely adopted metrics—accuracy and area under the receiver operating characteristic curve (AUROC), which quantifies threshold-independent discriminative ability. We found that augmenting real data with PhysDGM-generated samples consistently surpassed both the real-only baseline and mixtures with alternative generators (Fig~\ref{fig:figure4}i). In terms of accuracy, PhysDGM-augmented model achieved gains of $\uparrow$13\% over real-only, $\uparrow$117\% over mixtures of real and DiT, and $\uparrow$131\% over mixtures of real and DiffWave. For AUROC, the corresponding improvements were $\uparrow$4\%, $\uparrow$18\%, and $\uparrow$22\%, respectively.

To further characterize discrimination across decision thresholds, we plotted ROC curves for models trained on real data and on a mixture of real and PhysDGM synthetic data (Fig.~\ref{fig:figure4}m,n). After augmentation with PhysDGM, the ROC curve shifted upward and leftward, with the micro-average AUROC increasing from 0.955 to 0.972 ($\uparrow$1.7\%) and the macro-average AUROC from 0.943 to 0.967 ($\uparrow$2.5\%). These results indicate that PhysDGM enhances both sensitivity and overall accuracy, enabling more stable fault identification, earlier warning, and improved risk control in complex industrial environments.

\subsubsection{PhysDGM achieves full-data performance with 20× less training data}

We first assessed the capacity of PhysDGM to generate high-fidelity data under extreme data scarcity condition by training the model on a subset containing only 5\% of the turbofan engine RUL prediction dataset. Next, we systematically compared downstream prediction performance across three training configurations: (1) a low-data baseline using only 5\% original data, (2) a performance benchmark using 100\% complete data, and (3) a data-augmented set combining the 5\% original data with our generated synthetic data (Fig.~\ref{fig:figure4}o-r).

Remarkably, data augmentation restored prediction accuracy to levels comparable with the full dataset. Compared to the low-data baseline, the augmentation strategy yielded significant RMSE reductions, ranging from 29.97\% to 47.56\%. This finding indicates that PhysDGM can efficiently capture the intrinsic distribution and key features from extremely limited samples and generate high-fidelity data that effectively substitutes for large amounts of real data. This capability highlights its immense application potential in overcoming data scarcity and reducing acquisition costs.

\begin{figure*}[!th]
  \centering
  \includegraphics[width=0.96\linewidth]{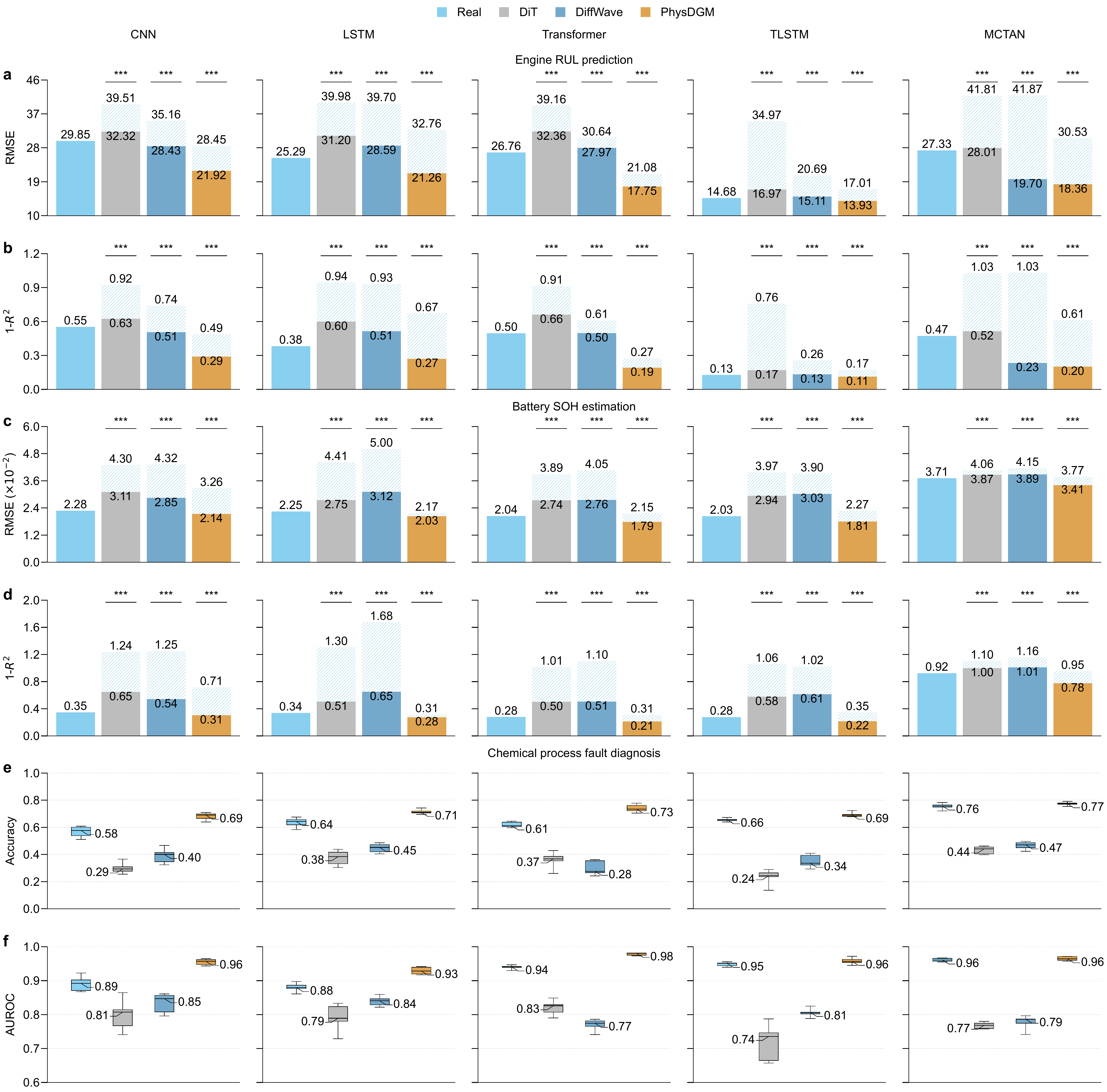}
  \caption{
    \textbf{Generalization of data generated by PhysDGM across tasks and models.}
    \textbf{a,b}, Engine RUL prediction with CNN, LSTM, Transformer, TLSTM, and MCTAN; models trained with PhysDGM-generated data achieve lower RMSE and 1-$R^2$ than those trained on raw data and on other synthetic mixtures. Under the conformal prediction (CP) framework, RMSE and $R^2$ is used as the nonconformity score to provide coverage guarantees.
    For statistical significance, we report one-sided Wilcoxon signed-rank test $P$ values across 1,000 independent runs (*** $P<0.001$, ** $P<0.01$, * $P<0.05$).
    \textbf{c,d}, Battery SOH estimation shows the same pattern, with lower RMSE and 1-$R^2$ for PhysDGM-based training than for the raw-data baseline and competing generators.
    \textbf{e,f}, Chemical process fault diagnosis evaluated by accuracy and AUROC; box plots summarize five independent experiments, with whiskers indicating the minimum and maximum, boxes spanning Q1 to Q3, and the central line denoting the median. PhysDGM yields higher scores and greater stability than the real-only setting and other synthetic-data mixtures. 
  }
  \label{fig:figure5}
\end{figure*}


\subsubsection{PhysDGM markedly enhances industrial prediction tasks across diverse models}
We show that adding PhysDGM-generated synthetic data to the training set significantly improves the performance of diverse downstream models on industrial prediction tasks. We use RMSE and the coefficient of determination ($R^2$) on the turbofan engine RUL prediction and battery SOH estimation datasets. The comparison includes models trained only on real data and models trained on a 1:1 mixture of real and synthetic data of different types.

We first evaluate PhysDGM on the RUL prediction dataset under multiple operating conditions, with the detailed results reported in Supplementary Data Tables~\ref{tab:fd01_cnn}--\ref{tab:fd04_trans}. We then compute performance metrics averaged over these operating conditions and present the comparison using bar charts (Fig.~\ref{fig:figure5}a,b). In terms of RMSE, replacing the original training data with synthetic data generated by PhysDGM reduces the RMSE of CNN by 7.93~($29.85 \to 21.92$), LSTM by 4.03~($25.29 \to 21.26$), Transformer by 9.01~($26.76 \to 17.75$), TLSTM by 0.75~($14.68 \to 13.93$), and MCTAN by 8.97~($27.33 \to 18.36$). In terms of $1-R^2$, the reductions are 0.26~($0.55 \to 0.29$) for CNN, 0.11~($0.38 \to 0.27$) for LSTM, 0.31~($0.50 \to 0.19$) for Transformer, 0.02~($0.13 \to 0.11$) for TLSTM, and 0.27~($0.47 \to 0.20$) for MCTAN. By contrast, mixing real data with synthetic data generated by DiT or DiffWave yields improvements in these metrics that are no more than 5\% of the gains achieved with PhysDGM.

To further examine the generalization of PhysDGM-augmented training, we evaluate SOH estimation under multiple battery discharge conditions and report average model performance (Fig.~\ref{fig:figure5}c,d). Consistent with the RUL task, combining real and PhysDGM-generated data yields over 13\% average improvement compared to the real-data baseline. Moreover, PhysDGM outperformed augmentation strategies based on DiT or DiffWave. These findings indicate that PhysDGM-generated data faithfully capture realistic industrial data patterns and are suitable for industrial prediction tasks. 


In addition, we evaluated classification accuracy and AUROC on a chemical fault diagnosis dataset. We compared models trained exclusively on real data with those trained on 1:1 mixtures of real and synthetic data generated by different methods. Results are summarized with box plots across multiple runs (Fig.~\ref{fig:figure5}e,f). In terms of accuracy, replacing the original training data with synthetic data generated by PhysDGM yields a maximum relative improvement of nearly 20\% across the evaluated models. In terms of AUROC, PhysDGM achieves a maximum relative gain of about 8\%. Introducing PhysDGM-generated synthetic data consistently improves both accuracy and AUROC, with gains that clearly surpass those achieved with DiT or DiffWave, which offer little to no benefit.

Together, these results suggest that PhysDGM-generated synthetic data consistently enhances fault diagnosis performance across multiple downstream models in diverse industrial settings and outperform advanced generative models. This consistent improvement further confirms the generalization of PhysDGM across diverse downstream scenarios.

\section{Discussion}

We introduce PhysDGM, a physics-embedded diffusion framework that fundamentally advances synthetic time-series data generation by treating physical laws as a first-class design principle enforced across the entire data generation trajectory. In safety-critical settings where data are scarce and failures are costly, conventional generative models often synthesize statistically and physically implausible time-series data, undermining their trustworthiness and utility.  PhysDGM addresses this barrier by generating synthetic data that is both high-fidelity to real-world distributions and rigorously consistent with the underlying physical constraints. Trained across 34 datasets spanning aerospace, batteries, and chemical processes, PhysDGM delivers physically-consistent synthetic data that enable downstream models to achieve higher accuracy with drastically less real data. Remarkably, downstream models maintain full-data predictive accuracy using merely 5\% of the original samples, reducing data collection costs by 20×. This off-the-shelf capability unlocks practical routes to robust predictive maintenance, rare-fault modeling, and digital twins, marking a significant step toward deploying AI in the world's most demanding industrial environments.

Detailed analyses reveal the mechanisms underlying PhysDGM’s performance advantages. The cornerstone of our approach is the stepwise physical embedding, which enforces physical constraints at every reverse-diffusion step. This materially outperforms conventional final output regularization, preventing the accumulation of intermediate violations and yielding superior structural fidelity. Our ablations show this method directly reduces physics violation rates (for degradation, multi-sensor coupling, etc.) and improves the structural fidelity of the generated data. This process is further enhanced by a dynamic training curriculum that progressively strengthens physical constraints, preventing premature convergence to an overly rigid solution. This allows it to first learn the rich statistical patterns of the data before being steered onto the manifold of physical reality. Finally, our framework’s flexibility is ensured by gradient-guided sampling, which allows a pre-trained model to adapt to new physical conditions—such as a change in operational load or working environment—without any retraining, a vital feature for dynamic industrial process.

Beyond methodological advances, PhysDGM holds significant implications for the practical application of industrial AI models in real-world settings. First, it alleviates pervasive data scarcity: models trained on PhysDGM-generated data using only 5\% of the real training set approach full-data baselines, with turbofan RUL prediction gains up to 40.8\% in representative settings. Second, PhysDGM enables robust rare-fault modeling by generating millions of physics-consistent degradation and failure trajectories, which expanded training coverage and improved RUL and fault diagnosis performance. Finally, PhysDGM supports privacy-preserving data sharing by creating synthetic datasets that retain key temporal–physical patterns of industrial systems while concealing sensitive operational information, achieving performance comparable to real data across aero-engine, battery, and chemical process tasks.

Despite these advances, our framework has several limitations that define the frontier for future work. First, our current constraint library is most effective for differentiable or smoothly approximable physical laws. Handling discrete events, non-smooth dynamics like actuator saturation, or complex control logic remains a challenge for gradient-based methods and may require hybrid approaches. Second, the computational cost of diffusion sampling, while mitigated by our training-free adaptation, can be a bottleneck for generating high-dimensional, large-scale time series. Future work on distilled samplers or latent diffusion for time series is essential for scalability. Finally, while the framework has been validated on representative datasets
(e.g., turbine engines, battery, and chemical process), its generalizability across diverse industries (e.g., petrochemicals, aviation) requires further exploration. Future research should aim to develop a foundational model capable of generating time series data across multi-domain and multi-device settings, thereby establishing its robustness and versatility in industrial applications.

\section{Method}
\subsection{Framework Overview}
PhysDGM is the framework we propose to generate time-series data that comply with physical laws. Its architecture is designed to deeply integrate physical constraints into both the training and sampling phases, and it is built upon three core components. The first is a stepwise physical embedding mechanism, which iteratively enforces physical laws at each reverse diffusion step to provide fine-grained supervision throughout the generation process. The second component is the dynamic physical constraint training (DPCT) strategy. This strategy resolves the conflict between learning the data distribution and adhering to physical principles by allowing the model to fit the data freely in the early stages before progressively increasing the constraint strength. The third component is the gradient-guided physical sampling method, a flexible technique that enables the model to adapt to new physical environments during the sampling process without retraining. The specifics of each component will be detailed in the following subsections.

\subsection{Mathematical Background of Diffusion Models}

Diffusion models comprise two key processes: the \textbf{forward diffusion process} and the \textbf{reverse denoising process}.

In the forward diffusion process, samples from the data distribution (\(x_0 \sim q(x)\)) are progressively corrupted by adding noise $\epsilon$, eventually transforming into standard Gaussian noise \(x_T \sim \mathcal{N}(0, I)\). This process is described by the Markov chain:
\begin{equation}
q(x_t|x_{t-1}) = \mathcal{N}(x_t; \sqrt{1 - \beta_t}x_{t-1}, \beta_t I),
\end{equation}
where $x_{1:T}$ represent the intermediate states in the forward diffusion process, \(\beta_t \in (0, 1)\) represents the noise level added at each step, and \(t\) denotes the diffusion step.

The reverse denoising process is designed to train a neural network to gradually restore Gaussian noise to its original data distribution. 
\begin{equation}
p_\theta(x_{t-1}|x_t) = \mathcal{N}(x_{t-1}; \mu_\theta(x_t, t), \Sigma_\theta(x_t, t)),
\end{equation}

In accordance with the diffusion model framework, we parameterize the reverse process as:
\begin{equation}
\boldsymbol{\mu}_\theta(\mathbf{x}_t,t)=\frac{1}{\sqrt{\hat{\alpha}_t}}\left(\mathbf{x}_t-\frac{\beta_t}{\sqrt{1-\hat{\alpha}_t}}\boldsymbol{\epsilon}_\theta(\mathbf{x}_t,t)\right)
\end{equation}
where \(\boldsymbol{\epsilon}_\theta(x_t, t)\) is a learned denoising function that predicts \(\epsilon\) using the neural network. The training loss function is expressed as:
\begin{equation}
L(x_0) = \sum_{t=1}^T \boldsymbol{E}_{q(x_t|x_0)} \left\| \epsilon - \epsilon_\theta(x_t, t|x_c) \right\|_2^2 
\end{equation}

This function is used to optimize the discrepancy between the estimated noise $\epsilon_\theta(x_t, t)$ and the original added noise $\epsilon$ in forward process.  Conditioning variables \(\boldsymbol{x_c}\) (such as labels or context) enable guided generation. 

\subsection{Stepwise Embedding of Physical Laws in the Generative Process}
Conventional diffusion models restore a target data distribution, $x_0$, from Gaussian noise, $x_T$, through a reverse denoising process. This process is typically driven by a neural network, $\epsilon_\theta(x_t, t)$, which predicts the noise added at each timestep $t$ and is optimized using the following loss function:
\begin{equation}
    L(x_0) = \sum_{t=1}^{T} \mathbb{E}_{q(x_t|x_0)} \left\| \epsilon - \epsilon_\theta(x_t, t|x_c) \right\|^2_2
\end{equation}
However, this standard paradigm focuses solely on learning the static data distribution and cannot guarantee that the intermediate states, or even the final sample, will adhere to real-world physical laws---a critical deficiency for industrial applications.

To address this challenge, we propose the stepwise physical embedding mechanism. Unlike conventional methods that apply constraints only at the final output stage, our core idea is to embed physical regularization at \textbf{every step of the entire reverse diffusion process}. This fine-grained enforcement prevents the accumulation of physical violations during generation, ensuring that the final samples possess high physical fidelity. Accordingly, we define the total training objective for PhysDGM as a composite loss function:
\begin{equation}
    L_{\theta,x_c} = \sum_{t=1}^{T} \mathbb{E}_{x_0 \sim D, \epsilon \sim \mathcal{N}(0,I)} \left[ L^{(t)}_{\text{data}} + L^{(t)}_{\text{phys}} \right]
\end{equation}
Here, $L^{(t)}_{\text{data}}$ is the standard data fidelity term, i.e., the noise prediction loss, which ensures alignment with the real data distribution:
\begin{equation}
    L^{(t)}_{\text{data}} = \left| \epsilon - \epsilon_\theta(x_t, t | x_c) \right|^2
\end{equation}
The term $L^{(t)}_{\text{phys}}$ is our physics-consistency regularizer, which penalizes violations of physical laws at each timestep $t$:
\begin{equation}
    L^{(t)}_{\text{phys}} = \Phi(t) \sum_{i=1}^{n} f_{p_i}(\hat{x}^{(t)}_0)
\end{equation}
where $f_{p_i}$ represents the $i$-th physical constraint function, $n$ is the total number of constraints, and $\Phi(t)$ is a dynamic weighting function (Detailed in \textbf{Methods} section \textbf{Dynamic Physical Constraint Training}). Since physical laws cannot be applied directly to noise $\epsilon$, we first use a reparameterization to compute the predicted "clean" data, $\hat{x}^{(t)}_0$, from the predicted noise $\epsilon_\theta(x_t, t)$, and then calculate the physical loss on $\hat{x}^{(t)}_0$.

We integrate four classes of critical industrial physical constraints: (1) \textbf{Degradation Physical Constraint}, which reflect the monotonic trends of performance decay; (2) \textbf{Multi-sensor Coupling Physical Constraint}, which capture the interdependencies among sensor readings; (3) \textbf{Range Physical Constraint}, which limit sensor outputs to valid operational ranges; and (4) \textbf{Discrete Physical Constraint}, which ensure that synthesized data for certain sensors are discrete rather than continuous. In short, four critical physical constraints are integrated into the diffusion model to ensure its adherence to real-world physical laws (See \textbf{Methods} section \textbf{Embedded Physical Constraints} for detailed information). These constraints are derived from:  (1) equipment degradation laws, which guide monotonic trends in performance metrics; (2) multi-sensor coupling laws, reflecting the interdependencies among sensor readings; (3) maximum and minimum value constraints, which limit sensor outputs to valid ranges; and (4) discrete constraints, which ensure that synthesized data values are discrete rather than continuous.

\subsection{Dynamic Physical Constraint Training}
During the early stages of diffusion model training, the generated time series often lack physical consistency. Imposing strong physical constraints too early can destabilize the generation process. To address this issue, we introduce a \textbf{training step-aware} Dynamic Physical Constraint Training (DPCT) approach, which progressively increases the intensity of physical constraints, optimizing the training process.

The motivation arises from the observation that during the initial stages of training, the model's outputs are still far from the true data manifold. Consequently, lower constraint intensities are required early to provide the model with greater flexibility to fit the data distribution. As training progresses, the model approximates the real data more closely. Stronger physical constraints are necessary to ensure that the synthesized data adhere to the physical laws. The DPCT is designed to meet the following requirements.

1) Constraint Adjustment: In the early stages of training, the intensity of physical constraints should remain low, allowing the model sufficient freedom to fit the data. As training advances, particularly in later stages, the intensity of the constraint should increase significantly to ensure the physical consistency of the generated data.

2) Gradual Transition: To maintain stability, the constraint intensity should increase gradually during the initial stages of training, avoiding premature restriction. In the later stages, the intensity should accelerate, driving the model toward solutions that conform to physical laws.

3) Boundary Conditions: At the start of training ($t = 0$), the constraint intensity should be minimal ($y = 0$). By the end of training ($t \to T$), the intensity should reach its maximum value ($y \to 1$).

Based on these considerations, we design the dynamic constraint adjustment function $\Phi(\cdot)$ as follows. Extended Data Fig.~\ref{fig:performance-summary-cptac} illustrates the curves of the method.
\begin{equation}
    y =\Phi(t)= \text{Sigmoid}\left(\alpha \cdot \ln\left(\frac{x}{1 - x}\right)\right), \quad x = \frac{t}{T}
\end{equation}
where \( y \) denotes the output constraint intensity, ranging between 0 and 1. The Sigmoid function smoothly maps the input to the range \([0, 1]\), with slow growth for small inputs and accelerated growth for larger inputs. The normalized training progress \( x = \frac{t}{T} \), where \( t \) is the current training step and \( T \) is the total number of training steps, quantifies the fraction of training completed. The term \( \alpha \cdot \ln\left(\frac{x}{1 - x}\right) \), a scaled logarithmic odds function, controls the rate and extent of change in the constraint intensity during training. The factor $\alpha$ determines the sharpness of the transition, with larger values resulting in steeper changes in intensity.

\subsection{Gradient-guided Physical Conditional Sampling}

The Gradient-guided Physical Conditional Sampling (GPCS) method is employed to generate sensor data with new physical constraints without requiring model retraining. Unlike traditional diffusion models, which necessitate retraining for each specific physical constraint, GPCS integrates physical constraints into pre-trained models using differentiable gradient guidance. This approach also incorporates conditional information, such as device labels and health status, during the sampling process to generate labeled time-series.

\begin{algorithm}[h]
\caption{Gradient-guided Physical Conditional Sampling}
\label{Gradient-guided-Physical}
\KwIn{The well-trained model $\theta$, diffusion total timestep $T$, noise schedules $\{\beta_t\}_{t=1}^T$, condition dataset $D_c$}
\KwOut{Industrial physics-constrain-synthetic multi-sensor time series}
Sample $x_T \sim \mathcal{N}(0, I)$\;
Obtain $x_c$ from condition dataset $D_c$\;
Physics-constraint loss function $f_{p_i}$\;

\For{$t = T, \ldots, 1$}{
    Sample $\epsilon \sim \mathcal{N}(0, I)$\;
    \If{$t = 0$}{
        $\epsilon = 0$\;
    }
Calculate latent variable $x_{t-1}$ according to Eqs.\;
$\hat{\epsilon}=\epsilon_\theta(x_t,t|xc)-\sqrt{1-\alpha_t}\cdot \sum_{i=1}^{n}\nabla_{x_t} f_{p_i}$\;
$x_{t-1} = \frac{1}{\sqrt{\alpha_t}} \left( x_t - \frac{\beta_t}{\sqrt{1 - \alpha_t}} \hat{\epsilon} \right) + \frac{1 - \bar{\alpha}_{t-1}}{1 - \bar{\alpha}_t} \beta_t \epsilon$\;
}
\Return{$x_0$}
\end{algorithm}

Firstly, as detailed in Algorithm \ref{Gradient-guided-Physical}, the reverse sampling process without incorporating physical information can be defined as:

\begin{equation}
x_{t-1} = \frac{1}{\sqrt{\alpha_{t}}}\left(x_{t} - \frac{\beta_{t}}{\sqrt{1 - \bar{\alpha}_{t}}}\epsilon_\theta(x_t, t | x_c)\right) + \frac{1 - \bar{\alpha}_{t-1}}{1 - \bar{\alpha}_t} \beta_t \epsilon\end{equation}
where \( x_t \) represents the data input to the denoising model at time step \( t \), \( \beta_t \) denotes the noise schedule, and \( \alpha_t = 1 - \beta_t \), with \( \bar{\alpha}_t = \prod_{i=1}^{t} \alpha_i \). Additionally, \( \hat{\epsilon} \) represents the noise estimation predicted by the denoising model.
This process iterates repeatedly until \( x_0 \), the generated sample, is obtained. The inclusion of conditional information in this process enables the generation of labeled industrial time series data. 

Then, the core idea behind the gradient adjustment strategy is to leverage gradient information to correct the initial noise prediction, ensuring that the generated samples more accurately reflect physical characteristics of the time-series. For new physical constraint loss \( f_{p_i} \), the model computes the gradient of \( f_{p_i} \) with respect to the input data \( x_t \), \( \nabla_{x_t} f_{p_i} \), to determine the direction for correcting the noise prediction:

\begin{equation} \nabla_{x_t} f_{p_i}=\frac{\partial f_{p_i}}{\partial {x_t}} \end{equation}

Subsequently, the model normalizes the gradient by \( \sqrt{1 - \alpha_t} \), and adjusts the noise prediction \( \epsilon_\theta(x_t, t | x_c) \) by subtracting the normalized gradient. This correction directs the noise prediction to more closely align with the physical constraints:

\begin{equation}\hat{\epsilon}=\epsilon_\theta(x_t,t|x_c)-\sqrt{1-\alpha_t}\cdot \sum_{i=1}^{n}\nabla_{x_t} f_{p_i}\end{equation}

The modified noise \( \hat{\epsilon} \) incorporates the multiple physical constraints $ \sum_{i=1}^{n}\nabla_{x_t} f_{p_i}$ of industrial sensor data. By adjusting the noise prediction without requiring additional training, this approach significantly reduces training costs and enhances the feasibility of practical applications.

\subsection{Embedded Physical Constraints}
\label{app:physcis}
To ensure that the generated time-series data conform to real-world physical principles, four representative and widely applicable physical laws are embedded into PhysDGM framework. These laws capture the essential dynamics commonly observed across diverse dynamical systems, including  degradation physical, multi-sensor coupling, range-bounded behaviors, and discrete operational states. Specifically,

(1) the Degradation Physical Law characterizes the monotonic deterioration trends of equipment performance, reflecting phenomena such as the gradual loss of turbine efficiency, bearing wear, or battery capacity fade during operation;

(2) the Coupling Physical Law describes the interdependence among correlated sensor signals governed by thermodynamic and fluid-mechanical relations, for example, the linkage between airflow, pressure, and rotational speed in aero-engines or between temperature and reaction rate in chemical processes;

(3) the Range Physical Law enforces the bounded domain of sensor measurements, ensuring that generated values remain within feasible physical limits, such as temperature, voltage, or rotational speed ranges defined by material and safety constraints; and

(4) the Discrete Physical Law reflects the quantized nature of operational transitions in control systems, such as valve opening and closing, pump on/off cycles, or gear position changes.

\subsubsection{Degradation Physical Law}
Industrial equipment often undergoes performance degradation over time due to factors such as wear and tear. For instance, NASA's report on engine failure modes highlights that as an engine approaches failure, certain performance metrics, such as compressor efficiency, tend to degrade. This degradation causes some of the sensor data to exhibit a monotonically increasing or decreasing trend, reflecting the progressive deterioration of the equipment's performance. The trend in these sensor readings can fluctuate as the engine ages, providing critical insight into the degradation process.
\begin{equation}
\omega_{\text{fan}}(t) = \omega_{\text{fan,0}} + k \cdot \Delta P_{\text{compressor}}
\end{equation}
where \(\omega_{\text{fan}}(t)\) is the fan speed at time \(t\), \(\omega_{\text{fan,0}}\) is the initial fan speed, and \(\Delta P_{\text{compressor}}\) is the pressure difference before and after the compressor. As the engine degrades over time, the performance of the compressor gradually declines, leading to a reduction in efficiency. This results in an increase in \(\Delta P_{\text{compressor}}\). 
Consequently, the physical fan speed and corrected fan speed will rise rapidly, while the cool air flow through the low-pressure turbines will decrease. These changes in sensor readings are indicative of the underlying physical degradation processes occurring within the engine.

\textbf{Constraints of monotonic increasing or decreasing trends}: To ensure the generated data adhere to realistic engine degradation processes, we introduce a regularization term based on a sliding average to enforce the monotonic properties described above. Specifically, let the sliding average of the generated data be \( D_i^{(t)} \). For data that are expected to grow monotonically, the first \( k \) terms of the sliding mean should be less than or equal to the last \( k \) terms of the sliding mean. Therefore, the difference between the two should be less than or equal to 0. We penalize by retaining the portion that does not conform to monotonic growth, i.e., the difference is greater than 0, through the ReLU function, recording it as monotonic loss. The loss of data that meets the expected growth trend is set to 0, indicating that there is no penalty. Monotonically decreasing loss is the same. This will achieve the effect of monotonic increase/decrease, in line with the equipment degradation process.
\begin{equation}
D_i^{(t)} = \frac{1}{k} \sum_{j=t}^{t+k-1} D_{i,j}
\end{equation}

\begin{equation}
L_{\text{Inc}} = \alpha_1 \cdot \text{ReLU}\left(D_i^{(0)} - D_i^{(t_0 - k)} \right)
\end{equation}

\begin{equation}
L_{\text{Dec}} = \alpha_2 \cdot \text{ReLU}\left(D_i^{(t_0 - k)} - D_i^{(0)} \right)
\end{equation}
where \( k \), \( i \), and \( t \) denote the size of the sliding window, the number of columns, and the time step, respectively. \( t_0 \) is the total time step. By constraining the first \( k \) items of the sliding mean \( D_i(0) \) of the generated data to be less than or equal to, or greater than or equal to, the last \( k \) items of the sliding mean \( D_i(t_0 - k) \), we ensure that the generated data follow the expected monotonic degradation trends. Additionally, by introducing gains \( \alpha_1 \) and \( \alpha_2 \), we encourage effective model training. This constraint-driven approach allows the model to capture the underlying degradation trends more accurately, ensuring generated data aligns more closely with the actual degradation processes observed in industrial equipment.

\subsubsection{Multi-sensor Coupling Physical Laws}

In industrial equipment, multi-sensor coupling relationships often exist, where measurements from certain sensors are interdependent, as described by Bernoulli's equation~\cite{clancy_aerodynamics_1975} and turbulence model~\cite{fakhari2023optimizing}. 

Bernoulli's equation describes the relationship between a fluid's speed and pressure. This relationship can be represented using the Bernoulli equation:
\begin{equation}
p + \frac{1}{2} \rho u^2 = p_T
\end{equation}
where \( p \) is the static pressure of the gas, \( \rho \) and \( u \) represent the gas density and the gas flow rate, respectively, and \( p_T \) is the total gas pressure. The total pressure of the gas is the sum of the gas static pressure \( p \) and the gas dynamic pressure \( \frac{1}{2} \rho u^2 \). In the intake of an aircraft engine, the total pressure \( P_2 \) at the fan inlet represents the total pressure of the gas \( p_T \), and the speed \( v_1 \) of the fan intake represents the flow rate of the gas \( u \). Therefore, the relationship between \( P_2 \) and \( v_1 \) can be expressed by the Bernoulli equation described above:
\begin{equation}
P_2 = p + \frac{1}{2} \rho v_1^2
\end{equation}
Meanwhile, the fan speed \( N_1 \) is a key factor influencing air compression efficiency. As \( N_1 \) increases, the intake air velocity \( v_1 \) rises, improving the fan's efficiency and increasing the total pressure \( P_2 \) at the fan inlet.
In all turbulence-modeled compressors of the turbulence SST \( k-\omega \) model, the inlet and outlet air pressures \( P_s \) and \( P_0 \) follow an isentropic compression relationship~\cite{peebles2023scalable}:
\begin{equation}
P_0 = P_s \times \left( 1 + \frac{\gamma - 1}{2} \text{Ma}^2 \right)^{\frac{\gamma}{\gamma - 1}}
\end{equation}
where \( \gamma \) is the specific heat ratio, \( R \) is the gas constant, \( \text{Ma} = \frac{V}{c} \), and \( c = \sqrt{\gamma R T_{\text{s}}} \) is the speed of sound. In an aircraft engine, the total pressure at the outlet of the fan and the total pressure at the inlet of the fan can be regarded as the outlet and inlet air pressures \( P_0 \) and \( P_s \). Therefore, the relationship between \( P_{21} \) and \( P_2 \) can be expressed by the above equation for isentropic compression:
\begin{equation}
P_{21} = P_2 \times \left( 1 + \frac{\gamma - 1}{2} \text{Ma}^2 \right)^{\frac{\gamma}{\gamma - 1}}
\end{equation}
where \( c = \sqrt{\gamma R T_{\text{in}}} \) in \( \text{Ma} = \frac{V}{c} \), and \( T_{\text{in}} \) is the inlet temperature. Absolute velocity \( V \) is positively correlated with fan speed \( N_1 \). This equation shows that increasing \( N_1 \) improves air compression and increases the Mach number \( \text{Ma} \) and \( P_{21} \). As \( P_{21} \) increases, the Low Pressure Compressor (LPC) obtains a higher pressure, which results in a greater compression ratio and leads to a corresponding increase in the Low Pressure Compressor Outlet \( P_{24} \).

In conclusion, the pressures at each stage of compression are interdependent. Specifically, the total pressures at the fan inlet (\( P_2 \)), fan outlet (\( P_{21} \)), LPC outlet (\( P_{24} \))
, as well as the physical fan speed (\( N_1 \)), show synchronous changes.

\textbf{Multi-sensor Coupling Constraints}:
To capture the trend of synchronous changes, this paper introduces a consistent slope loss function. This function ensures that the trends of different sensors remain aligned by comparing their slope variations.

Assume that \( x_i(t) \) denotes the value of the \( i \)-th feature in the time series data at moment \( t \), and express the slope between neighboring time points of the feature with the following equation:
\begin{equation}
\text{slope}(x_i) = \frac{x_i(t+1) - x_i(t)}{x_i(t) + \epsilon}
\end{equation}

\begin{equation}
L_{\text{slope}} = \frac{1}{n} \sum_{i} \text{MSE}(\text{slope}(x_i), \text{slope}(x_{\text{base}}))
\end{equation}
where \(\epsilon\) is a minimal value that prevents the denominator from being zero. Using the slope formula, we calculate the rate of change of each eigenvalue with respect to the previous time step of that eigenvalue, which represents the magnitude of change of the time series between two consecutive time points. These are denoted as the slopes of the five features mentioned earlier. The slope \( x_{\text{base}} \) of the Physical fan speed is used as the benchmark data, and the mean square error (MSE) is applied to quantify the difference between the slopes of the benchmark features and the slopes of the other features. Finally, by minimizing the error between the slopes of each feature and the baseline feature, we ensure that the trend of different features in the time dimension remains consistent.
\subsubsection{Range Physical Law}

Industrial time series usually have a fixed range of values. As shown in the figure, the sensor values for the fan-corrected speed have two tiny ranges between which the values fluctuate. This occurs because the corrected fan speed $\text{NR}_f$ is automatically adjusted by the control system, switching between high power mode ($\text{NR}_f \approx 1$) and low power mode ($\text{NR}_f \approx 0$). The sensor value shows two modes of maximum and minimum values.

Generative models need to ensure that the numerical range of the output data is consistent with the original data. The range of the minimum value after max-min normalization is $[0, k_\text{LB}]$, and the range of the maximum value is $[k_\text{UB}, 1]$. Therefore, we penalize synthetic data that exceed this bound. It should comply with the following physical constraints.
\begin{align}
L_\text{UB} &= \frac{1}{n} \sum_{i=1}^n \text{ReLU}(x_i - k_\text{UB}) \\
L_\text{LB} &= \frac{1}{n} \sum_{i=1}^n \text{ReLU}(k_\text{LB} - x_i) \\
L_\text{realm} &= \alpha \sum_{i=1}^n (L_\text{LB} + L_\text{UB})
\end{align}
where $k_\text{UB}$ and $k_\text{LB}$ denote the values of upper and lower boundaries respectively. If the generated data exceed the limit, the difference between it and the limit will be greater than 0. In contrast, the difference between the data inside the boundary and the boundary will be less than or equal to 0. The part of the difference that is greater than 0 is retained by the ReLU function and penalized as the loss for exceeding the boundary. And the parts less than or equal to zero are all set to zero, representing no loss. The final loss function is the sum $L_\text{realm}$ of all the above boundary penalization terms, where $\alpha$ serves as the weighting coefficient. In this way, all sensor data are generated within the desired range, which improves the stability and accuracy of the model.

\subsubsection{Discrete Physical Law}

Time-series data typically consists of two types of values: continuous values, which vary over a wide range and can take any real value within a given physical domain, and discrete values, which are limited to a finite set of predefined values that correspond to specific states, events, or conditions. For example, the Bleed Enthalpy is normalized to be uniformly discretized within the range of 0 to 1.
To effectively model discretely distributed data, we introduce a discrete loss function that penalizes generated data that does not align with the predefined discrete values, ensuring that the model correctly learns and preserves the discrete characteristics inherent in the sensor data.
\begin{equation}
    X_{\text{disc}} = X \cdot k_{\text{disc}} \quad \% \, 1
\end{equation}
\begin{equation}
    L_{\text{disc}} = \frac{\alpha}{n} \sum_{i=1}^{n} \left( \sin \left( X_{\text{disc}} \cdot 2\pi - \frac{\pi}{2} \right) + 1 \right) \cdot \frac{1}{2}
\end{equation}
where \( n \) represents the number of data points, and \( k_{\text{disc}} \) denotes the number of discrete values in the original data. Ideally, the generated data, when expanded by a factor of \( k_{\text{disc}} \), should yield integer values corresponding to the discrete values. However, if the generated data do not meet this expectation, it will contain fractional values. To quantify this deviation from being an integer, we apply the modulus operation to isolate the fractional part of the generated data.
When the generated data deviates from the intended discrete values, the fractional part will differ significantly from 0 or 1. To penalize such deviations, we introduce a sine-based loss function. This function increases in magnitude as the fractional part moves farther from 0 or 1, thereby assigning a larger loss value to more significant deviations. This approach effectively guides the model to generate data that adheres to the expected discrete distribution.

\textbf{Physical Constraints of Lithium Battery Degradation}:
For the generation of the battery SOH estimation dataset, we develop a physics-informed neural network (PINN) framework~\cite{wang2024physics} to model lithium-ion battery degradation characteristics. Given the inherent complexity of electrochemical degradation processes, which resist explicit mathematical formulation, we propose a hybrid architecture that synergistically integrates data-driven deep learning with physics-based constraints. The model employs a multilayer perceptron (MLP) as its core feature extractor, processing multidimensional battery system features \( \mathbf{X} \in \mathbb{R}^d \) through nonlinear transformations \( \Phi: \mathbf{X} \to \mathbf{u} \) to predict battery health states \( \mathbf{u} \), with corresponding state-of-health (SOH) labels \( \mathbf{y} \) for each battery group.

To incorporate domain knowledge, we introduce a novel physics-constrained loss function that regularizes the degradation trajectory:

\begin{equation}
\text{Loss}_{\text{deg}} = \sum_{i=1}^{B-1} \text{ReLU} \left( (\mathbf{u}_{i+1} - \mathbf{u}_i) \cdot (\mathbf{y}_i - \mathbf{y}_{i+1}) \right)
\end{equation}
where \( B \) represents batch size, and \( \mathbf{u}_i \), \( \mathbf{u}_{i+1} \), \( \mathbf{y}_i \), \( \mathbf{y}_{i+1} \) denote predicted and actual health states at consecutive time steps. This formulation ensures physical consistency: when predicted and actual degradation trends align, the loss term vanishes due to the ReLU activation; conversely, conflicting trends incur positive penalties. This dual-objective optimization simultaneously enforces data fidelity and electrochemical consistency.

The physics-informed training paradigm guarantees that learned health state transitions align with empirical degradation patterns, yielding synthetic battery data with enhanced physical plausibility and practical utility. This approach effectively bridges the gap between data-driven modeling and fundamental electrochemical principles, addressing the challenge of insufficient data for battery health prediction.

\subsection{Experiment Setup}
\label{app:exp}

For the engine systems, four categories of physical constraints are incorporated: (1) multi-sensor coupling constraints influenced by aerodynamic principles, (2) degradation constraints derived from equipment failure patterns,  (3) range constraints specifying valid sensor output boundaries, and (4) discrete constraints ensuring that synthesized values match the discrete nature of certain sensor readings. On the lithium battery datasets, degradation constraints are embedded to guide the generation process, as they capture the characteristic aging patterns of lithium-ion batteries. 

\subsubsection{Dataset Description}
\label{sec:dataset}
The dataset for turbofan engine RUL prediction contains degradation trajectories from 259 turbofan engines. It is divided into four subsets (FD01-FD04) corresponding to distinct failure modes and operational conditions. The data features 21 sensor readings capturing engine health, with 14 utilized after preprocessing. A detailed description is provided in Supplementary Data Table~\ref{tab:C-MAPSS}.

The dataset for aero-engine HI prediction reflects real-world operational conditions, capturing the full degradation process from a healthy state to failure. It includes key operating parameters such as altitude, speed, and power. We utilize the DS02 sub-dataset, characterized by high dimensionality, complex operating conditions, and diverse degradation modes. A detailed description is provided in Supplementary Data Tables~\ref{tab:N-CMAPSS} and~\ref{tab:Description of N-CMAPSS}.

The dataset for battery SOH estimation~\cite{wang2024physics} contains 17 key performance indicators (KPIs) and State of Health data from charging and discharging experiments on 55 lithium-ion batteries. The experiments were conducted at Xi'an Jiaotong University and include six distinct protocols, covering fixed-current charging, random-walk modes, and GEO satellite operation simulations, providing a foundation for analyzing degradation under diverse operational conditions (See Supplementary Data Table~\ref{tab:battery_dataset_description}).

The dataset for chemical process fault diagnosis~\cite{downs1993plant} is a simulation of a plant-wide industrial chemical process based on an actual industrial plant. This process simulates a reactor/separator/recycle arrangement involving gas-liquid exothermic reactions to produce two products from four reactants. It involves eight components and four irreversible reactions, providing 41 measurements and 12 manipulated variables. The simulation, developed for diagnostics, encompasses 20 programmed disturbances (faults), which we summarized into 14 categories (Supplementary Data Table~\ref{tab:TEP_Disturbances}).

\subsubsection{Baseline and Implementation Details}
We selected multiple baseline methods for comparative analysis to comprehensively evaluate the performance of the proposed generative models. These methods include PINN~\cite{wang2024physics} and diffusion modeling methods such as DiffWave~\cite{kong2022diffwave}, SSSD~\cite{lopezalcaraz2022diffusionbased}, TabDDPM~\cite{kotelnikov2023tabddpm}, DiT~\cite{peebles2023scalable}, and Diff-TS~\cite{yuandiffusion}. For each of these methods, we conducted multiple rounds of training on the two datasets and selected their best performance as the final test result.

In our experiments, the model was trained for 70 epochs on NVIDIA RTX 3090 GPUs using a linear noise schedule with 500 time steps ranging from 0.0001 to 0.02. The initial learning rate was set to 2 $\times 10^{-3}$. Data generation used the DDPM framework with a 500-step backward sampling strategy.
The architecture follows a U-Net design, with the encoder comprising four modules, progressively increasing channels by [1, 2, 4, 8] using 4×4 convolutional kernels with a stride of 2. The decoder restores the shape through transposed convolutions, with each module incorporating two SiLU-activated convolutional layers and residual connections for enhanced feature representation. A detailed description for PhysDGM is provided in Supplementary Data Table~\ref{tab:parameters}.

\subsubsection{Quality Evaluations Metrics}
\label{sec:metrics}

In order to fully assess the quality of the generated data in fidelity, utility, and diversity, the following metrics are used:

\textbf{Discriminative Score (DS): }  
This metric measures fidelity, i.e., how indistinguishable the generated data is from the real data. A classifier with a two-layer RNN is employed to differentiate between real and generated data, and its classification accuracy is calculated. The absolute difference between the classification accuracy and 0.5 is used as the  discriminative score; the lower the score, the higher the similarity between the generated data and the real data.  

\textbf{Predictive Score (PS): }  
This metric evaluates data utility, i.e., how well the generated data performs in downstream prediction tasks. To assess this, we train a Transformer model using only the generated data and then evaluate it on the original test data. The performance indicators are task-specific. For degradation prediction tasks, we calculate the RMSE, where lower values indicate higher utility. For fault diagnosis tasks, we measure classification accuracy, where higher values indicate higher utility. 

\textbf{Visualization: }  
This metric assesses diversity by ensuring that the distribution of the generated data covers the real data distribution. The t-SNE dimensionality reduction method is applied to project high-dimensional features into low-dimensional space, enabling intuitive evaluation of the data quality through visualization.

\section*{Data Availability}
The turbofan engine RUL prediction data and corresponding labels are available from the NASA data portal (\url{https://data.nasa.gov/dataset/cmapss-jet-engine-simulated-data}). The aero-engine HI prediction data and corresponding labels are available from the NASA Intelligent Systems Division (\url{https://www.nasa.gov/intelligent-systems-division/}). In case the link does not work, the dataset can also be accessed via \url{https://drive.google.com/drive/folders/1UzUwcq_I9RD7NOr6sI6S2Je14gDUxdK6?usp=sharing}. The battery SOH estimation data and corresponding labels are available from the public repository maintained by Fujin Wang (\url{https://wang-fujin.github.io/}). The chemical process fault diagnosis data and corresponding labels are available from the Tennessee Eastman Process repository (\url{https://github.com/camaramm/tennessee-eastman-profBraatz}).

\section*{Code availability}
All code was implemented in Python using PyTorch as the primary deep-learning library. The complete pipeline for data processing, model training, evaluating is available at \url{https://github.com/Dolphin-wang/PhysDGM} and can be used to reproduce all experiments presented in this paper.

\section*{Acknowledgments}
The research is supported by the NSFC (National Science Foundation of China) project No.62225302, 623B2014, 62173023.

\section*{Author contributions statement}
H.W. conceived the study and designed the experiments. H.W., Y.L., and T.W. developed the methodology, wrote the codes, performed the experiments, and analyzed the experimental results. Y.Z. and D.J. assisted in data collection and validation. H.W., Y.L., T.W., and Y.Z. wrote the original manuscript. L.R. and X.Z. reviewed and revised the manuscript. L.R. supervised the project.  All authors contributed to manuscript preparation and approved the final version.
\bibliography{sn-bibliography}

\clearpage

\captionsetup[figure]{labelformat=empty}
\setcounter{figure}{0}
\captionsetup[table]{labelformat=empty}

\clearpage

\begin{appendices}
\label{sec:Appendix}

\begin{figure}[!ht]
    \centering
    \includegraphics[width=0.95\linewidth]{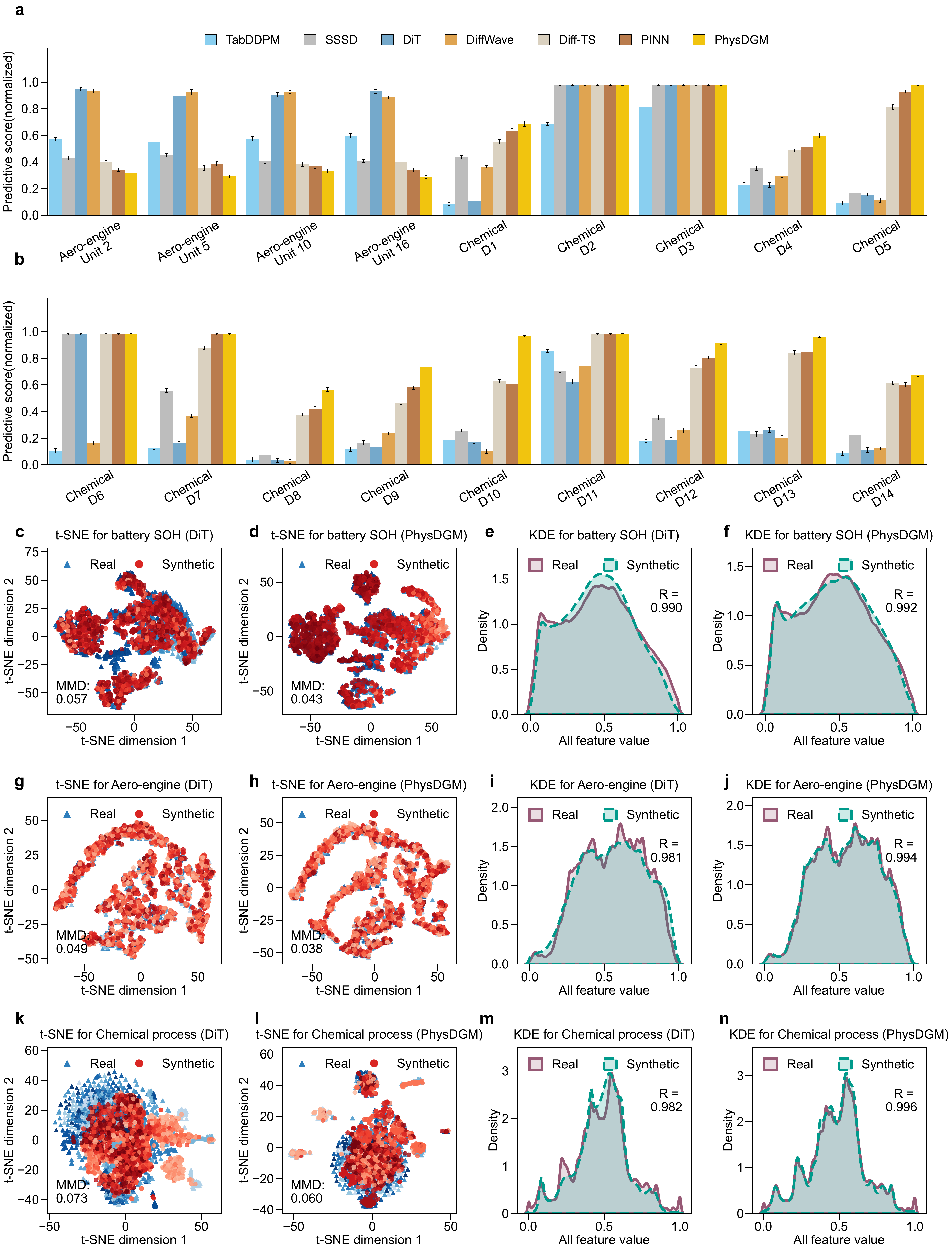}
    \caption{\textbf{Extended Data Fig 1}: \textbf{Quality inspection of data generated by PhysDGM.} \textbf{a,b}, Sub-task performance of the usability test for data generated by PhysDGM and baseline models on the chemical process fault diagnosis task, where classifiers trained on synthetic data are evaluated on real test samples using classification accuracy; the task comprises 14 classes, with class grouping details provided in the corresponding table. The mean values and standard deviations shown in a and b are derived from 7 independently trained models, each evaluated over 1000 experimental trials.
    \textbf{c-n},Comparison of real and generated data distributions for PhysDGM and DiT across three tasks---aero-engine HI prediction, battery SOH estimation, and chemical process fault diagnosis---using t-SNE and KDE visualizations. The t-SNE embeddings illustrate distributional overlap between real and synthetic samples, with the Maximum Mean Discrepancy (MMD) reported as a quantitative reference. The KDE plots depict probability density curves of key features, where the coefficient of determination ($R^2$) evaluates the goodness of fit between real and generated distributions.}
    \label{fig:exfig1}
\end{figure}

\clearpage

\begin{figure}[!ht]
    \centering
    \includegraphics[width=0.96\linewidth]{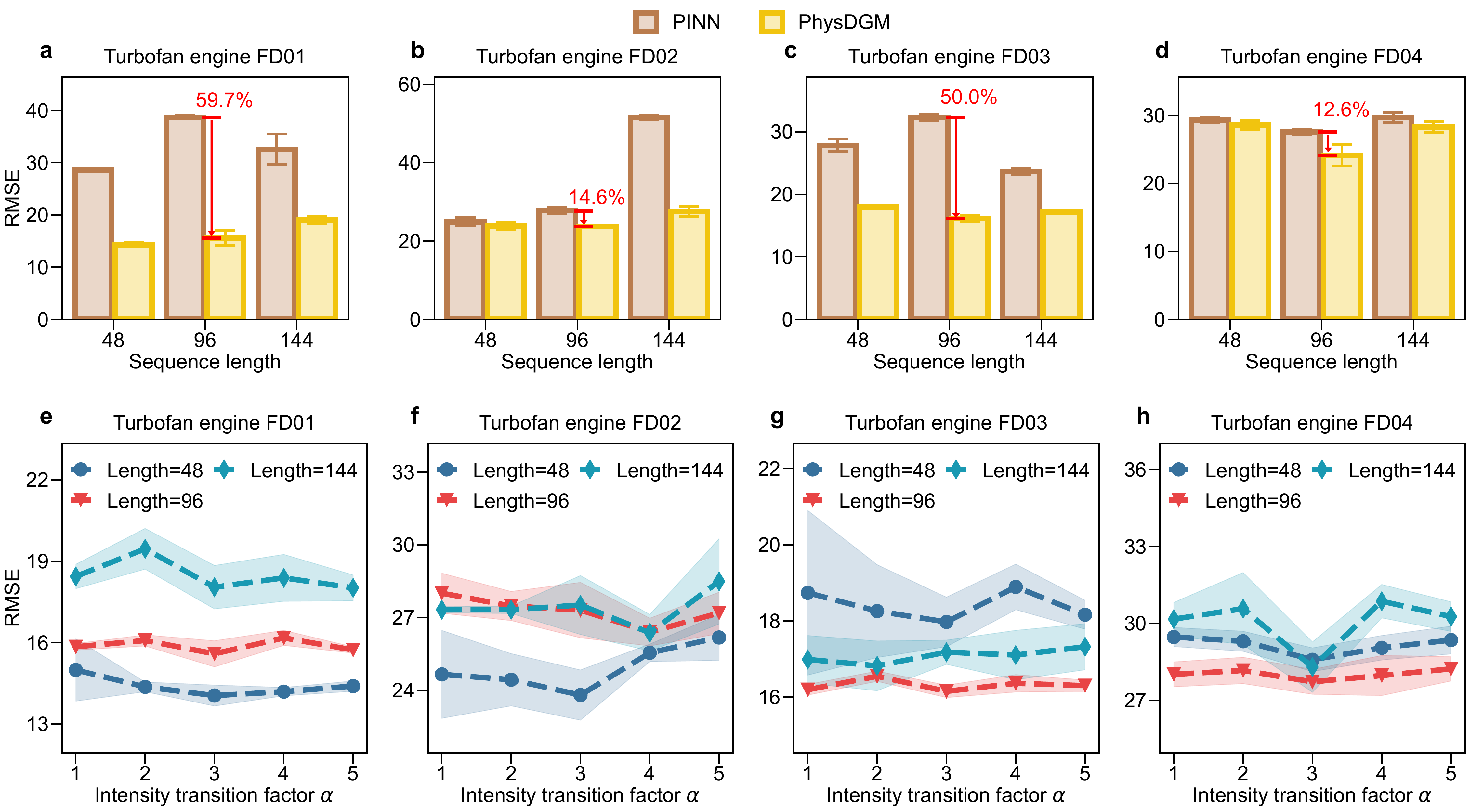}
    \caption{\textbf{Extended Data Fig 2}: \textbf{Effectiveness of dynamic physical constraint training (DPCT).} \textbf{a-d}, Performance comparison of PhysDGM and PINN on the turbofan engine RUL prediction task. The evaluation assesses performance based on prediction scores. The results demonstrate that PhysDGM consistently outperforms the PINN baseline across all tested operating conditions. The mean values and standard deviations shown in a--d are derived from 2 independently trained models, each evaluated over 1000 experimental trials. \textbf{e-h}, Parameter analysis of the constraint intensity transition factor $\alpha$. This plot illustrates the \textbf{Root Mean Square Error (RMSE)} of the predictive task for $\alpha$ values set to 1, 2, 3, 4, and 5. The analysis identifies $\alpha = 3$ as the optimal setting, ensuring a smooth yet effective enforcement of physical constraints throughout the training process.}
    \label{fig:exfig2}
\end{figure}

\clearpage

\begin{figure}[!ht]
    \centering
    \includegraphics[width=0.96\linewidth]{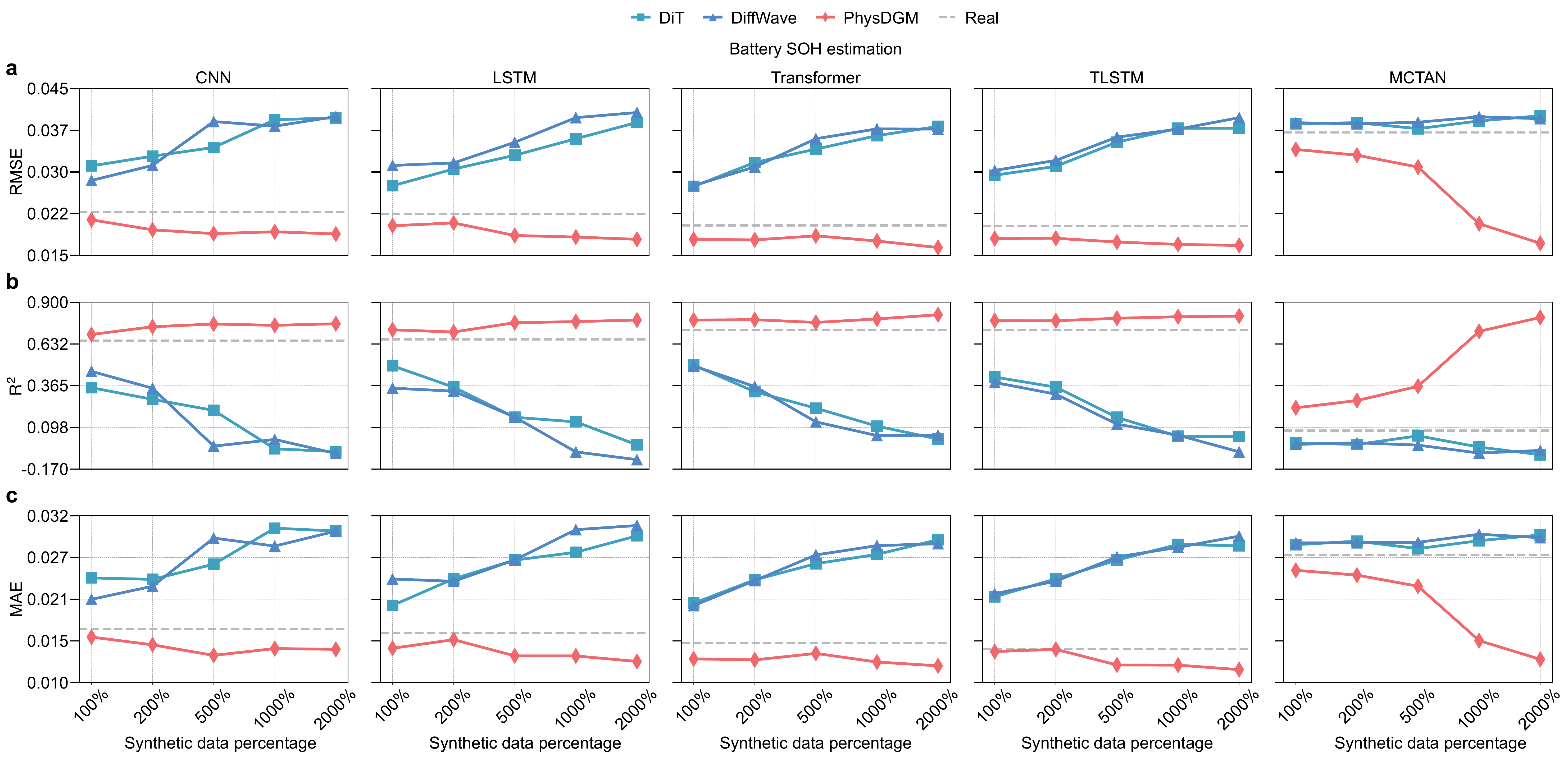}
    \caption{\textbf{Extended Data Fig 3}: \textbf{The generalization of data generated by PhysDGM across tasks and representative model architectures.} \textbf{a.b.c}, Extended figure illustrating battery SOH prediction using CNN, LSTM, Transformer, TLSTM, and MCTAN, where synthetic data with varying ratios is combined with original data to create mixed training datasets; model performance on specified test sets is quantified using RMSE, MAE, and $R^2$, showing that models trained with PhysDGM-generated data achieve lower errors and higher $R^2$ than those trained on raw data or other synthetic mixtures under the conformal prediction framework.}
    \label{fig:exfig3}
\end{figure}

\clearpage

\begin{figure}[!ht]
    \centering
    \includegraphics[width=0.6\linewidth]{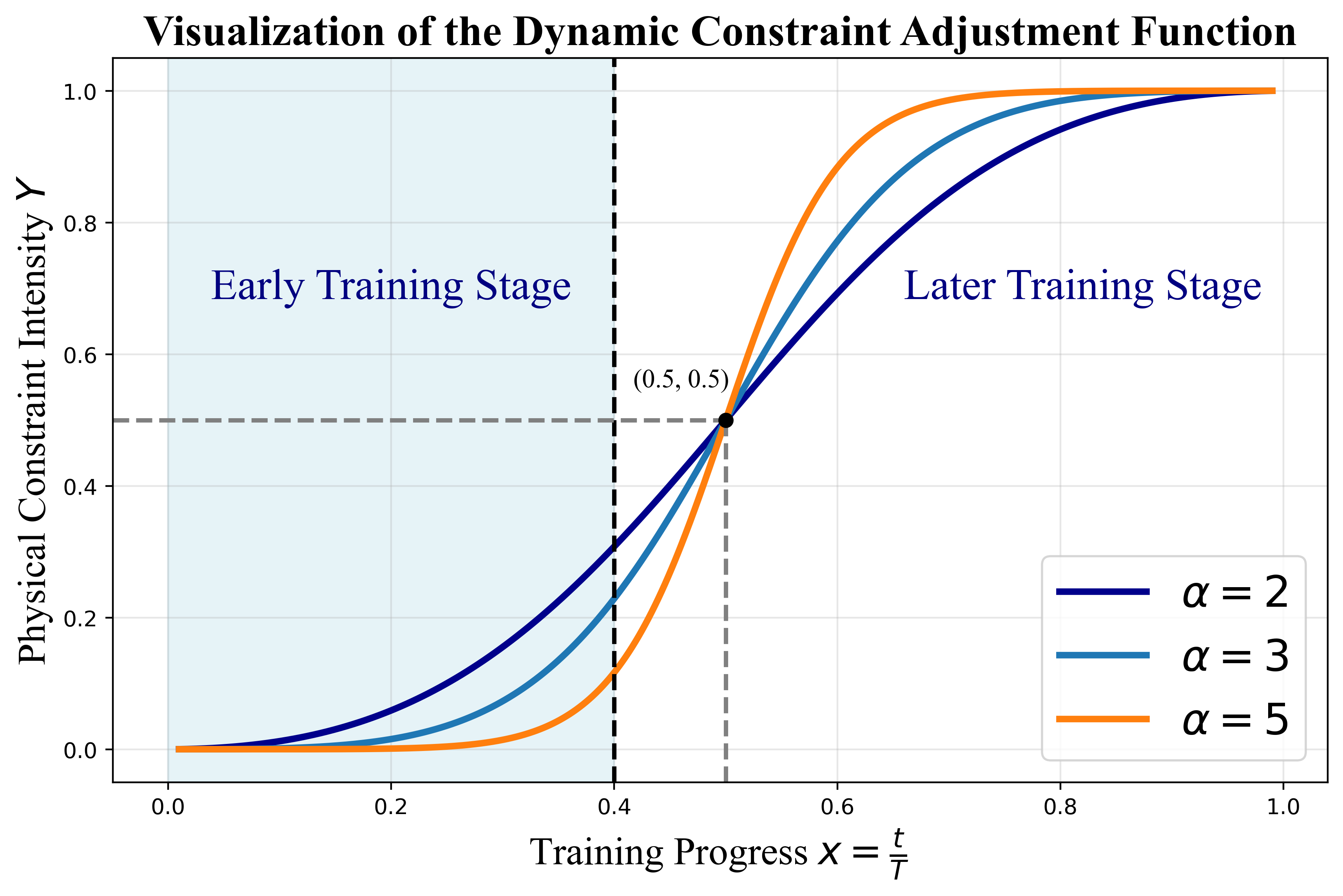}
    \caption{\textbf{Extended Data Fig 4}: \textbf{Visualization of the dynamic constraint adjustment function used in the training step-aware approach}. The curve represents the evolution of constraint intensity \(y\) as a function of the normalized training progress \(x = \frac{t}{T}\). Early in the training process, the constraint intensity remains low, allowing the model flexibility, while in the later stages, the intensity increases significantly to enforce physical consistency.}  
    \label{fig:performance-summary-cptac}
\end{figure}

\end{appendices}

\end{document}